\documentclass{article}

\makeatletter
\def\input@path{{./}}
\makeatother
\usepackage[main,final]{neurips_2026}
\usepackage{graphicx}
\usepackage[percent]{overpic}
\graphicspath{{./}}

\usepackage[utf8]{inputenc}
\usepackage[T1]{fontenc}
\usepackage{hyperref}
\usepackage{url}
\usepackage{booktabs}
\usepackage{array}
\usepackage{amsmath}
\usepackage{amsfonts}
\usepackage{nicefrac}
\usepackage{microtype}
\usepackage[dvipsnames]{xcolor}

\usepackage{multirow}

\newcommand{\opsd}{OPSD}
\newcommand{\qwenonesevenb}{Qwen3-1.7B}
\newcommand{\qwenfourb}{Qwen3-4B}
\newcommand{\qweneightb}{Qwen3-8B}
\newcommand{\qwenfourbinstruct}{Qwen3-4B-Instruct-2507}
\newcommand{\qwenfourbthink}{Qwen3-4B-Think-2507}
\newcommand{\olmosevenbinstruct}{OLMo-7B-Instruct}
\newcommand{\olmosevenbthink}{OLMo-7B-Think}
\title{Rethinking On-Policy Self-Distillation for Thinking Models}

\author{%
    Simran Kaur \quad
    Narutatsu Ri \quad
    Yinghui He \quad
    Liam Fowl \quad
    Sanjeev Arora \\
    Princeton Language and Intelligence (PLI)\\
    Princeton University\\
    \texttt{\{skaur,nr3764,yh0068,lf2728,arora\}@princeton.edu} \\
}

\begin{document}

\raggedbottom

\maketitle

\begin{abstract}

On-policy self-distillation (OPSD) is a promising approach to recursive self-improvement \citep{zhao2026selfdistilledreasoneronpolicyselfdistillation,shenfeld2026self,hubotter2026reinforcement}. In this setting, a model acts as its own teacher when augmented with privileged information (e.g. a solution to a math problem). Thinking models seem especially well positioned to teach themselves this way, since they can deliberate, verify, and self-correct. Surprisingly, we show that privileged self-distillation can make thinking models worse with respect to long reasoning traces. Across five Qwen3 and OLMo thinking models evaluated on AIME24, AIME25, and HMMT25, privileged-context distillation causes a relative drop of up to 17\% in avg@16 accuracy.
The degradation scales with the amount of privileged context withheld from the student and is most pronounced at long rollout budgets, where thinking models otherwise obtain their largest gains.
%This failure mode is not specific to self-distillation: on-policy distillation (OPD) improves thinking models, but privileged OPD reverses these gains.
% This failure mode is not specific to self-distillation: on-policy distillation (OPD) improves thinking models, but privileged OPD reverses these gains.
This failure mode extends to distillation from a larger, privileged teacher: on-policy distillation (OPD) improves thinking models, but privileged OPD reverses these gains.
%Our diagnostics suggest that this failure mode is linked to
%how privileged teacher context reshapes learning at high-entropy forking positions
Our diagnostics suggest that privileged teacher context reshapes learning at high-entropy forking positions
\citep{bigelow2024forking, zhang2026embarrassingly}, i.e., rollout positions where multiple continuations remain plausible and may lead to different reasoning paths.
Privileged context lowers fork rates in thinking-model rollouts but not in instruction-tuned model rollouts,
which may explain the observation that OPSD training helps instruct models more reliably than thinking models.
%This leads to an interesting dichotomy wherein
%This may explain an interesting performance dichotomy, where
%privileged context can help instruction-tuned models while hurting more performant thinking models that depend on exploration and rollout quality.
%This effect is especially visible at
At self-correction branches,
%where
privileged OPD penalizes sampled reconsideration tokens that vanilla OPD supports.
%This effect is especially visible when the student begins a self-correction branch, where privileged OPD penalizes sampled reconsideration tokens that vanilla OPD supports.
Consistent with this, thinking models trained with a privileged teacher produce fewer verification, backtracking, and hedging markers, even after length normalization.
These findings indicate that applying self-distillation methods to strong thinking models requires further consideration of token-level signal---especially around tokens related to correction and crucial reasoning steps.%
\footnote{Code: \url{https://github.com/princeton-pli/rethinking-opsd-for-thinking-models}}

\end{abstract}

\section{Introduction}
\label{sec:intro}

On-policy self-distillation (\opsd{}) has emerged as an exciting approach for self-improvement in language models \mbox{\citep{zhao2026selfdistilledreasoneronpolicyselfdistillation,shenfeld2026self,hubotter2026reinforcement}}.
In this setting, a single model plays the role of both a student and  teacher.
The teacher is provided additional privileged information, such as a gold solution, a final answer, or environmental feedback.

Thinking models are natural candidates for self-improvement. Post-trained to deliberate at test time \mbox{\citep{openai2024learning,deepseek2025r1,yang2025qwen3}},
they can branch into cases, verify intermediate steps, hedge, backtrack, and recover from errors before committing to an answer \mbox{\citep{arora2025training,gandhi2025cognitive,venhoff2025understanding}}.
Yet existing methods have mostly been studied outside this regime, using instruction-tuned models, short generation budgets, or trained on rollouts with thinking disabled  ~\citep{zhao2026selfdistilledreasoneronpolicyselfdistillation, shenfeld2026self, hubotter2026reinforcement}. This leaves open whether privileged-context self-distillation remains beneficial when the supervised trajectory is itself the long deliberation trace that thinking models rely on at test time.

In this paper, we report that existing privileged-context on-policy self-distillation methods can degrade thinking models, particularly at long rollout budgets. This degradation is not explained by the short training completion budget alone. Under the same short-budget on-policy training setup, vanilla OPD with a larger teacher improves the thinking student, whereas providing the teacher with privileged context reverses the gain. Our analysis suggests a plausible explanation, namely that privileged context may suppress the deliberative behaviors that thinking models rely on at test time.

We establish this failure mode through five observations. First, \opsd{} helps instruction-tuned models more reliably than thinking models. Second, under full-solution privileged context, \opsd{} degrades five OpenThoughts-trained thinking models across Qwen3 and OLMo. Third, the harm scales with how much privileged context the teacher receives. Final-answer-only context causes milder damage, while full-solution context causes substantially more. Fourth, the gold-demo-conditioned teacher itself does not show the same degradation. It produces much shorter rollouts while preserving high pass@k, whereas the trained student inherits shorter responses without the same longer-budget benefits. Fifth, the degradation is strongest at long rollout budgets and coincides with reduced fork rates, weaker self-correction signal, and fewer deliberation markers in trained students.

Such findings called for mechanistic exploration at the token level. The explanation appears to be a crucial sign reversal in the per-token distillation signal at high-entropy decision points, suggesting that access to privileged context suppresses exploration --e.g.,  branching, reconsideration, and uncertainty-marking moves. Thinking-model rollouts contain many such positions, which we call {\em forking positions} in line with recent work~\citep{bigelow2024forking,lin2024critical,vassoyan2025ignore,zhang2026embarrassingly}. They are often marked lexically by tokens such as \emph{wait}, \emph{hmm}, \emph{but}, and \emph{maybe}. Under vanilla on-policy distillation, these tokens carry positive advantage. Once privileged context is added, their advantage flips negative, and the trained student produces fewer of them at evaluation, even after length normalization.

We refer to this phenomenon as fork suppression, whereby privileged-context self-distillation undermines the very deliberative behaviors that made thinking models natural candidates for self-improvement in the first place.

Section~\ref{sec:method} describes the experimental setup; Section~\ref{sec:results} establishes the empirical pattern; Section~\ref{sec:analysis} analyzes the per-token signal at forking positions; Section~\ref{sec:relatedwork} situates the result among recent self-distillation methods; Section~\ref{sec:Discussion} discusses implications for self-improvement of thinking models.

\section{Experimental Setup}
\label{sec:method}

We evaluate whether privileged-context on-policy self-distillation preserves the
test-time search behavior of thinking models. We write the on-policy
distillation objective in a context-explicit form, since our experiments vary
precisely what additional information the teacher receives. Given an input
problem $x$, the student policy $\pi_S$ samples an on-policy rollout
$y \sim \pi_S(\cdot \mid x)$. We then train the student by minimizing a
token-level divergence between the teacher and student next-token distributions
along this sampled rollout:
\[
\mathcal{L}_{\mathrm{OPD}}
=
\sum_t
D\left(
\pi_T(\cdot \mid x, c, y_{<t}),
\pi_S(\cdot \mid x, y_{<t})
\right).
\]
Here $y_{<t}$ is the prefix before token $t$, $\pi_T$ is the teacher policy,
$\pi_S$ is the student policy being updated, and $c$ denotes teacher-only
privileged context. The divergence $D$ can be instantiated as forward KL,
reverse KL, JSD, or another token-level distillation divergence. Vanilla OPD
corresponds to $c=\emptyset$, while privileged-context OPD sets $c$ to
additional information such as a final answer or gold demonstration. In
\opsd{}, the teacher and student share the same architecture and typically start
from the same checkpoint; in our experiments, the teacher is initialized as a
copy of the initial student checkpoint and receives privileged context while the
student does not. Unless otherwise noted, training rollouts are capped at
4,096 completion tokens. We use JSD distillation for one epoch with effective
batch size 64; this training cap is shared by the vanilla OPD, privileged OPD,
and \opsd{} comparisons. Full hyperparameter details are in
Appendix~\ref{app:experimental-details}.

\paragraph{Training data.}
We use two training domains. For math reasoning, we train on an OpenThoughts
15k subset, using the problem as the student prompt and the full reference
solution as the privileged teacher context. For Countdown, we train on 15k
examples from \texttt{jasonrqh/Countdown-CoT-20k}; the privileged context is the
post-thinking solution suffix. The Countdown split also reserves 500 held-out
examples for in-domain evaluation.

\paragraph{Models and comparisons.}
Our main comparisons separate instruct models from thinking models. On
Countdown, we train paired instruct and thinking models from the Qwen3-4B
\citep{yang2025qwen3} and OLMo-3-7B \citep{olmo2025olmo3} families to test
whether \opsd{} behaves differently when the base model already performs
explicit deliberation. On OpenThoughts, we focus on thinking models across
sizes and families, including \qwenonesevenb{},
\qwenfourb{}, \qweneightb{}, \qwenfourbthink{}, and \olmosevenbthink{}. We also
include controls that compare \opsd{} to vanilla OPD and that disable thinking
during \opsd{} training while keeping thinking enabled at evaluation.

\paragraph{Evaluation.}
We evaluate transfer to AIME 2024, AIME 2025, and HMMT 2025, and evaluate
Countdown-trained models on the held-out Countdown split. For AIME/HMMT, each
benchmark has 30 problems. To reflect standard evaluation practice for thinking
models, which often uses long rollout budgets to benefit from test-time compute
scaling, our main evaluation adopts a long generation cap. We generate 16
samples per problem, each with a maximum length of 38,912 tokens. This
evaluation cap is much larger than the default 4,096-token training completion
cap. Main tables report avg@16 accuracy, the mean sample correctness over the
16 rollouts for each problem.
Note that avg@16 is distinct from unbiased
pass@16, which measures whether at least one of the 16 samples solves the
problem. To compare the main long-budget setting with shorter evaluation settings, we repeat evaluation at generation caps of 4,096, 8,192, 16,384, 32,768, and 38,912 tokens, and also report pass@1\footnote{Empirical pass@1 is the same per-sample correctness quantity as avg@16, estimated from sampled rollouts under the evaluation decoding distribution.} and unbiased pass@16.

\section{Results}
\label{sec:results}

\subsection{\opsd{} helps instruct models but can degrade thinking models}

\begin{table*}[t]
\centering
\caption{\textbf{\opsd{} degrades thinking models across model families.}
We train each thinking model with privileged OpenThoughts solution context and evaluate on AIME24, AIME25, and HMMT25. Entries report avg@16 accuracy, and the Average column averages the three benchmarks. \opsd{} lowers the average score for all five thinking models, suggesting that the degradation is not specific to one model size or family.}
\label{tab:openthoughts_thinking_models}
\small
\setlength{\tabcolsep}{5pt}
\renewcommand{\arraystretch}{1.15}

\newcommand{\metrichead}[1]{%
  \textbf{#1}%
}

\begin{tabular*}{\textwidth}{@{\extracolsep{\fill}}llcccc@{}}
\toprule
\textbf{Model}
&
& \metrichead{AIME24}
& \metrichead{AIME25}
& \metrichead{HMMT25}
& \metrichead{Average} \\
\midrule

\multirow{2}{*}{\textbf{\qwenonesevenb{} (Thinking)}}
& \textbf{Base}
& \textbf{0.502}
& \textbf{0.398}
& \textbf{0.215}
& \textbf{0.372} \\
&
\textbf{+\opsd{}}
& 0.435
& 0.302
& 0.185
& 0.308 \\
\midrule

\multirow{2}{*}{\textbf{\qwenfourb{} (Thinking)}}
& \textbf{Base}
& \textbf{0.727}
& \textbf{0.635}
& \textbf{0.410}
& \textbf{0.591} \\
&
\textbf{+\opsd{}}
& 0.683
& 0.556
& 0.362
& 0.534 \\
\midrule

\multirow{2}{*}{\textbf{\qweneightb{} (Thinking)}}
& \textbf{Base}
& \textbf{0.758}
& \textbf{0.700}
& \textbf{0.448}
& \textbf{0.635} \\
&
\textbf{+\opsd{}}
& 0.721
& 0.613
& 0.400
& 0.578 \\
\midrule

\multirow{2}{*}{\textbf{\qwenfourbthink{}}}
& \textbf{Base}
& \textbf{0.804}
& \textbf{0.804}
& \textbf{0.552}
& \textbf{0.720} \\
&
\textbf{+\opsd{}}
& 0.787
& 0.731
& 0.529
& 0.683 \\
\midrule

\multirow{2}{*}{\textbf{\olmosevenbthink{}}}
& \textbf{Base}
& \textbf{0.719}
& \textbf{0.667}
& \textbf{0.452}
& \textbf{0.612} \\
&
\textbf{+\opsd{}}
& 0.715
& 0.652
& 0.446
& 0.604 \\

\bottomrule
\end{tabular*}
\end{table*}

\opsd{} training helps instruct models more reliably than
thinking models. In the Countdown-trained setting
(Table~\ref{tab:countdown_think_vs_instruct}), \qwenfourbinstruct{} and
\olmosevenbinstruct{} improve under \opsd{} ($0.553 \to 0.559$ and
$0.488 \to 0.494$ avg@16 performance). The matched thinking models show mixed
results. \qwenfourbthink{} drops slightly ($0.776 \to 0.765$), while
\olmosevenbthink{} gains modestly ($0.679 \to 0.695$).
Appendix~\ref{app:sd-zero-srt} shows a similar pattern in an SD-Zero-style
self-revision pipeline \citep{he2026selfdistillationzero}. The self-revision
stage helps both the instruction-tuned and thinking models, but the subsequent
\opsd{} stage further helps only the instruction-tuned model and hurts the
thinking model.

Thinking model degradation is clearest in the OpenThoughts-trained setting
(Table~\ref{tab:openthoughts_thinking_models}). Across model scale and family,
all five thinking models we evaluate lose avg@16 performance under the same
training recipe, with drops ranging from $6.4$ points (\qwenonesevenb{}) to
$0.8$ points (\olmosevenbthink{}).
Repeating \opsd{} training across three seeds yields the same pattern,
with every run scoring below its corresponding base model (Appendix~\ref{app:training-seed-robustness}). Paired problem-level bootstrap intervals for the main average deltas are
reported in Appendix~\ref{app:bootstrap-cis}.

\begin{table}[t]
\centering
\caption{\textbf{The degradation depends on whether thinking is enabled during \opsd{} training.}
We train on OpenThoughts 30k and evaluate with thinking mode in all rows. When thinking is enabled during \opsd{} training, \qwenfourb{} loses avg@16 accuracy on all three math evaluations; disabling thinking during training preserves the base-model average while still evaluating with thinking enabled.}
\label{tab:openthoughts_thinking_enabled_disabled}
\small
\setlength{\tabcolsep}{8pt}
\renewcommand{\arraystretch}{1.15}
\begin{tabular*}{\linewidth}{@{\extracolsep{\fill}}lcccc@{}}
\toprule
\textbf{Method}
& \textbf{AIME24}
& \textbf{AIME25}
& \textbf{HMMT25}
& \textbf{Avg.} \\
\midrule
\qwenfourb{}
& 0.727
& 0.635
& 0.410
& 0.591 \\
+\opsd{} (Thinking)
& 0.667
& 0.571
& 0.358
& 0.532 \\
+\opsd{} (No Thinking)
& 0.731
& 0.615
& 0.423
& 0.590 \\
\bottomrule
\end{tabular*}
\end{table}

For thinking models, \opsd{} degrades performance when the rollouts being supervised are thinking rollouts.
Table~\ref{tab:openthoughts_thinking_enabled_disabled} holds
the model fixed (\qwenfourb{}) and evaluates with thinking enabled in all rows.
When training rollouts are non-thinking, performance is preserved ($0.591 \to 0.590$); when they are thinking, performance drops to $0.532$.

\subsection{Thinking-model degradation is specific to teacher-side privileged context}

\begin{table*}[t]
\centering
\caption{\textbf{Privileged teacher context, not on-policy distillation itself, degrades thinking-model performance.}
On a \qwenonesevenb{} thinking student, vanilla OPD with a larger \qweneightb{} teacher improves avg@16 accuracy over the base model. However, conditioning the teacher on information unavailable to the student at test time makes the resulting on-policy update harmful on average. This occurs both for a larger teacher (OPD gold demo) and for the self-teacher (OPSD) setting. All OPD/OPSD variants are trained with a 4,096-token completion cap; all methods, including the base model, are evaluated with a 38,912-token generation cap.}
\label{tab:qwen17_openthoughts_opsd_opd}
\footnotesize
\setlength{\tabcolsep}{3pt}
\renewcommand{\arraystretch}{1.12}

\newcommand{\metrichead}[1]{%
  \textbf{#1}%
}

\begin{tabular*}{\textwidth}{@{\extracolsep{\fill}}llcccc@{}}
\toprule
\textbf{Model}
&
& \metrichead{AIME24}
& \metrichead{AIME25}
& \metrichead{HMMT25}
& \metrichead{Average} \\
\midrule

\multirow{4}{*}{\textbf{\qwenonesevenb{}}}
& \textbf{Base}
& 0.502
& \textbf{0.398}
& 0.215
& 0.372 \\

&
\textbf{+\opsd{}}
& 0.435
& 0.302
& 0.185
& 0.308 \\

&
\textbf{+OPD}
& \textbf{0.540}
& 0.385
& \textbf{0.252}
& \textbf{0.392} \\

&
\textbf{+OPD gold demo}
& 0.467
& 0.344
& 0.240
& 0.350 \\

\bottomrule
\end{tabular*}
\end{table*}

The same student improves under vanilla on-policy distillation, where the teacher is
not given privileged context. In
Table~\ref{tab:qwen17_openthoughts_opsd_opd}, vanilla OPD trains
\qwenonesevenb{} against a larger teacher that sees the same prompt, improving
avg@16 performance from $0.372$ to $0.392$. Adding privileged context reverses
the sign. Context-enhanced OPD with a privileged gold demonstration reduces
performance to $0.350$, and \opsd{} with a full gold demonstration reduces it
to $0.308$. Performance falls when
the teacher scores the student's trajectory while conditioned on information the
student will not have at test time.

\subsection{The degradation appears when models are allowed to think longer}

\begin{figure}[t]
    \centering
    \includegraphics[width=\linewidth]{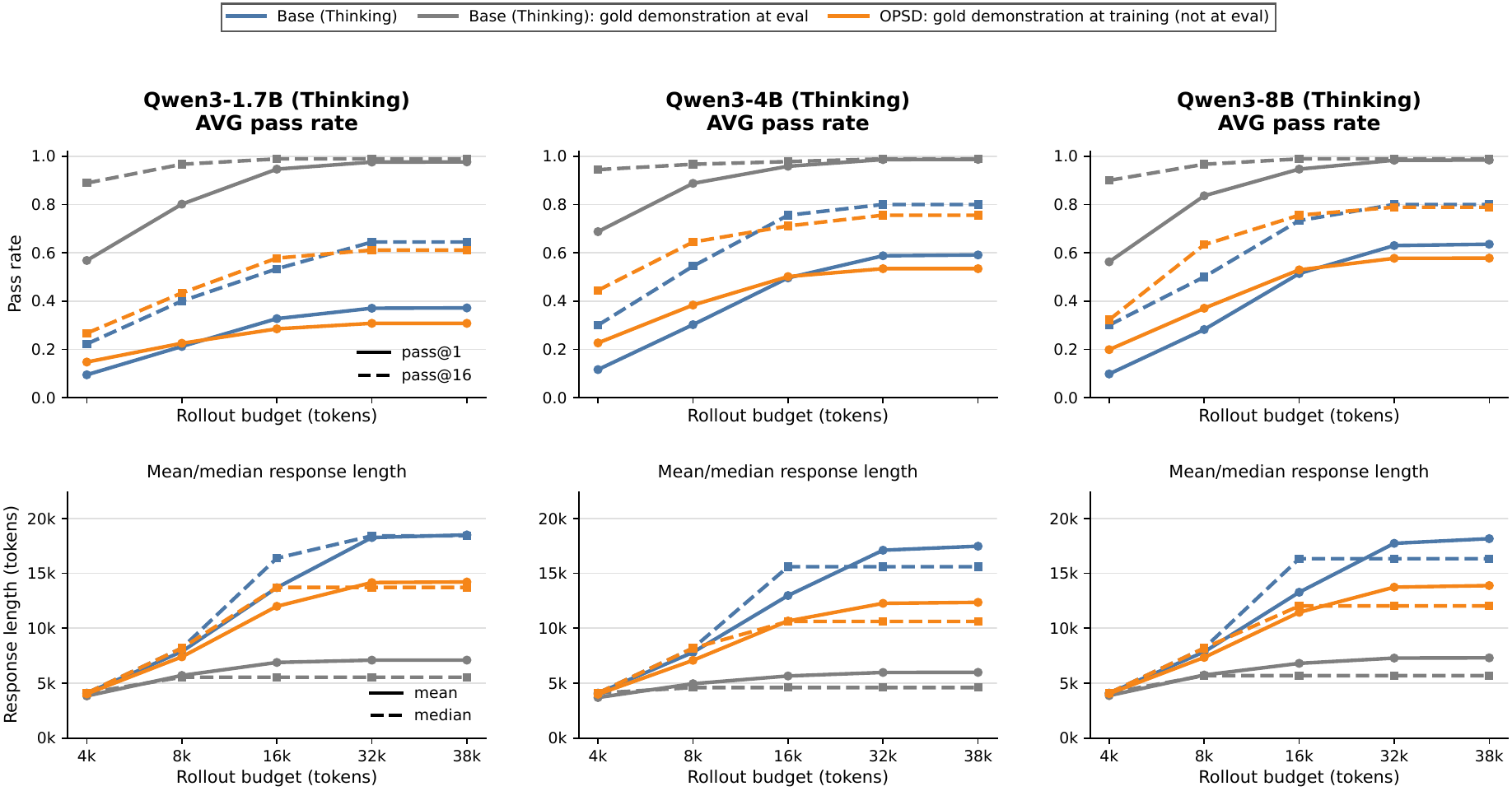}
    \caption{
    \textbf{For thinking models, \opsd{} can improve short-budget performance via compression but can hurt long-budget reasoning.}
    We evaluate OpenThoughts-trained Qwen3-4B, Qwen3-8B, and OLMo-3-7B thinking models across rollout budgets. Models are trained with a 4,096-token completion cap and evaluated at generation caps from 4,096 to 38,912 tokens. Top row: pass@1 and pass@16, averaged over AIME24, AIME25, and HMMT25. Bottom row: mean and median response length. \opsd{} with gold demonstrations often helps at 4k--8k tokens, but the advantage shrinks or reverses at 32k--38k tokens, where responses become substantially shorter than the base model.
    }
    \label{fig:openthoughts_accuracy_length_collapse}
\end{figure}

The degradation in thinking-model performance is concentrated at long rollout
budgets.
Figure~\ref{fig:openthoughts_accuracy_length_collapse} evaluates
OpenThoughts-trained \qwenfourb{}, \qweneightb{}, and \olmosevenbthink{} across
rollout budgets from 4k to 38k tokens. At 4k--8k tokens, \opsd{} models perform
comparably to or above their bases\footnote{Throughout this section, ``base''
denotes the corresponding instruct or thinking checkpoint before additional
\opsd{}/OPD training, not a pretrained base model.}. By 32k--38k tokens, they match or fall
below. Response length follows the same pattern. At small budgets, base and
\opsd{} rollouts are similar in length; at large budgets, \opsd{} rollouts are
substantially shorter. Overall, \opsd{} removes the gains that thinking models obtain from longer rollouts.

One might suspect a simpler explanation based on the mismatch between training
and evaluation lengths. Training rollouts are capped at 4,096 tokens, whereas
evaluation allows much longer rollouts, so short-budget training may cause the
model to forget longer reasoning behaviors. The OPD experiments rule out this
explanation as the sole cause. In
Table~\ref{tab:qwen17_openthoughts_opsd_opd}, vanilla OPD with a larger
unprivileged teacher improves the student under the same short-budget on-policy
setup, whereas adding teacher-only gold-demonstration context reverses the gain.
The corresponding budget curves in
Figure~\ref{fig:qwen17_opd_gold_demo_budget_curves} show that the two OPD
variants differ not only in accuracy but also in realized response length.
Vanilla OPD largely preserves the base response-length behavior, while
gold-demonstration OPD shortens responses and reduces the pass@1 gains. Thus,
the drop cannot be attributed only to the train/eval budget mismatch; it also
depends on whether the teacher scores the short-budget rollout with privileged
context.

\subsection{Students inherit shorter rollouts but not the teacher's pass@k gains}

We next evaluate whether the gold-demo-conditioned base model---the \opsd{}
teacher setup---exhibits the long-budget degradation observed in the \opsd{}
student. Figure~\ref{fig:openthoughts_accuracy_length_collapse} compares three
settings for each Qwen3 thinking-model size: the base model without privileged
context, the \opsd{} teacher (the base model when provided the in-context gold
demonstration), and the \opsd{}-trained student evaluated without the in-context
gold demonstration. The teacher's own rollouts become much shorter, but its
pass@k remains high and often improves. The student also produces shorter
responses, but compared with the teacher's gains over
the unprivileged base, the student's pass@k gains are smaller at short budgets
and disappear or reverse at longer budgets. Thus, the student inherits the
teacher's shorter response behavior more reliably than the teacher's
privileged-context performance benefits.

\subsection{Sparse privileged context preserves long-budget behavior better than dense privileged context}

\begin{figure}[t]
    \centering
    \includegraphics[width=\linewidth]{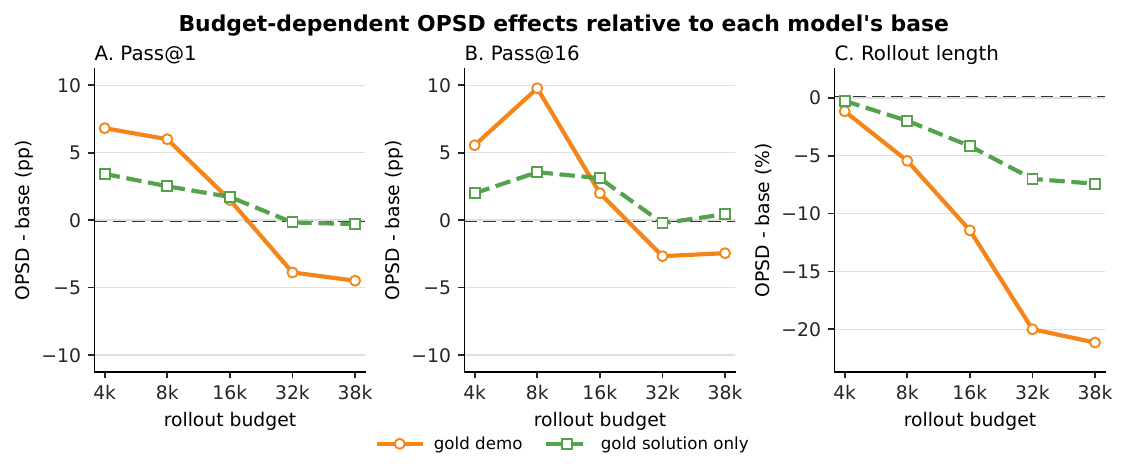}
    \caption{
    \textbf{In thinking models, sparse privileged context preserves long-budget behavior better than dense demonstrations.}
    We plot \opsd{} minus each model's base performance across rollout budgets, averaged over the thinking models in Table~\ref{tab:openthoughts_thinking_models}. Panels A and B show pass@1 and pass@16 changes in percentage points; Panel C shows relative rollout-length change. Dense gold demonstrations give larger short-budget gains but turn negative at long budgets and strongly shorten responses, while the sparse gold-solution hint remains closer to the base model.
    }
    \label{fig:openthoughts_sparse_context}
\end{figure}

We next ask whether the student-side degradation depends on how much privileged
context the teacher receives, and observe that full gold demonstrations degrade
long-budget behavior more than final-answer-only context.
Figure~\ref{fig:openthoughts_sparse_context} compares these two forms
of privileged context, averaged over the models in
Table~\ref{tab:openthoughts_thinking_models}. Full demonstrations give the
largest short-budget gains in pass@1 and pass@16, but also the largest
long-budget reversals, with mean rollout length compressed to roughly
$0.8\times$ the base at 38k tokens. The final-answer-only condition gives
smaller short-budget gains, remains closer to the base at long budgets, and
produces less length compression. Because the answer-only and full-demonstration
runs use the same training recipe and 4,096-token completion cap, this difference
cannot be explained by the short training budget alone; it depends on how much
privileged context the teacher receives.

In Section~\ref{sec:analysis}, we show that \opsd{} reduces fork rates at
high-entropy decision points and reduces explicit deliberation markers in
evaluation rollouts.

\section{Analysis: Privileged Context Reduces Forking}
\label{sec:analysis}

Section~\ref{sec:results} isolates when privileged-context distillation damages
thinking models. The degradation appears most clearly at long rollout budgets
and grows with the density of the teacher's privileged context. We next analyze
a possible mechanism. Privileged context changes the per-token distillation
signal on forking tokens, which mark uncertainty or redirection, and trained
students produce fewer such tokens at evaluation. The evidence has three
parts. First, privileged context shifts the fork--lock structure of
thinking-model rollouts
(\S\ref{sec:fork-lock-analysis}). Second, at these positions, the token-level
distillation signal reverses on self-correction and uncertainty markers
(\S\ref{sec:token-signal-analysis}). Third, trained \opsd{} students produce
fewer deliberation markers at evaluation
(\S\ref{sec:trained-student-deliberation}).

\subsection{Privileged context shifts the fork--lock distribution}
\label{sec:fork-lock-analysis}

Following the SSD diagnostic of \citet{zhang2026embarrassingly}, we distinguish
\emph{fork} positions from \emph{lock} positions. A fork is a high-entropy
position where several plausible continuations remain available and may lead to
different reasoning paths; a lock is a locally constrained position where the
continuation is comparatively determined. We classify positions from the
teacher distribution along sampled rollouts with a top-candidate dominance
heuristic: a position is a fork when the top candidate has probability below
$0.25$, and a lock when the top candidate has probability above $0.65$. We then
report per-rollout fork and lock rates.

\begin{figure}[t]
    \centering
    \begin{overpic}[width=\linewidth]{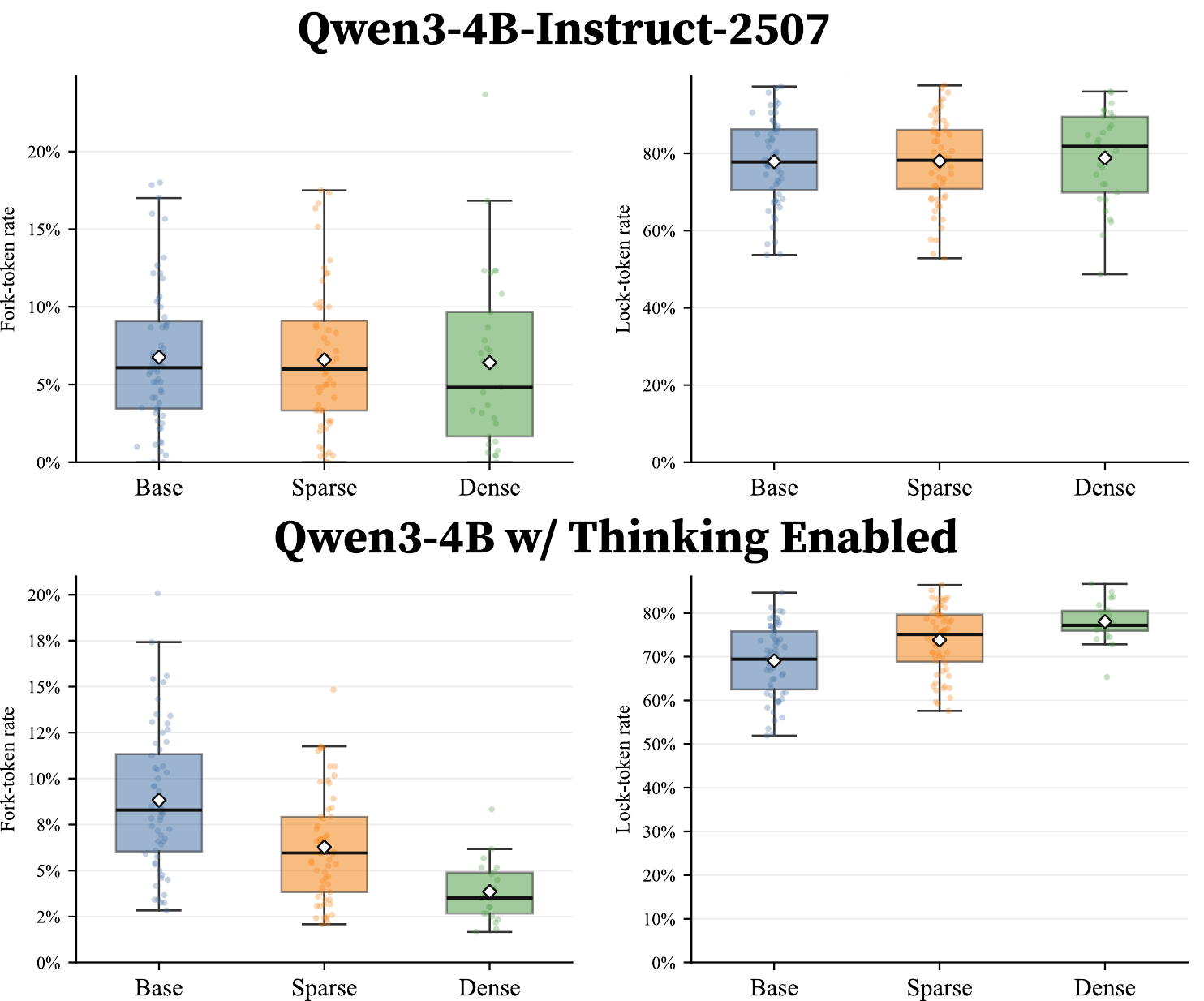}
        \put(21,77.4){\color{white}\rule{58\unitlength}{6.5\unitlength}}
        \put(50,80.4){\makebox(0,0){\large\bfseries \qwenfourbinstruct{}}}
        \put(19,35.6){\color{white}\rule{62\unitlength}{6.5\unitlength}}
        \put(50,38.6){\makebox(0,0){\large\bfseries Qwen3-4B w/ Thinking Enabled}}
    \end{overpic}
    \caption{
    \textbf{Dense privileged context lowers fork rates in thinking rollouts but has little effect on instruction-style rollouts.}
    We compute fork and lock rates using the SSD-style diagnostic of \citet{zhang2026embarrassingly}: fork positions are high-entropy decision points with multiple plausible continuations, while lock positions are locally determined continuations. Panels show the \opsd{} side of the diagnostic for base, sparse privileged-context, and dense privileged-context conditions; the left column shows fork rate and the right column shows lock rate. For the instruction-tuned model, fork and lock rates remain mostly flat as privileged context becomes denser, with at most a small dense-context reduction in fork rate. For the thinking model, denser privileged context lowers fork rate and raises lock rate, consistent with the hypothesis that \opsd{} suppresses the branching points needed for test-time search.
    }
    \label{fig:fork_lock_context}
\end{figure}

Figure~\ref{fig:fork_lock_context} compares \qwenfourbinstruct{} and
\qwenfourb{} with thinking enabled under three teacher conditions: no
privileged context, sparse context containing the gold final answer, and dense
context containing the full gold demonstration. The two modes respond
differently. For the instruction-tuned model, fork rates remain near $0.06$ and
lock rates near $0.78$, essentially flat across conditions. For the thinking
model, denser privileged context monotonically lowers fork rates and raises
lock rates. The median fork rate drops from roughly $0.083$ to
$0.035$ between the base and dense conditions, while the median lock rate rises
from roughly $0.69$ to $0.77$.

Privileged context reshapes the rollout structure of thinking models but leaves
instruction-style rollouts essentially unchanged.

\subsection{Privileged context reverses the signal on fork markers}
\label{sec:token-signal-analysis}

The fork-rate changes in \S\ref{sec:fork-lock-analysis} align with a sign
reversal in the per-token distillation signal. Forks are often marked by short
epistemic and revision tokens such as \textit{wait}, \textit{hmm},
\textit{but}, and \textit{maybe}. These markers are lexical proxies for forking
positions, not direct measurements. We inspect the per-token teacher--student
log-ratio along sampled rollouts: positive values mean the teacher assigns more
probability than the student to the sampled token, and negative values mean the
teacher assigns less.

\begin{figure}[t]
    \centering
    \includegraphics[width=0.95\linewidth,trim=0 270 0 0,clip]{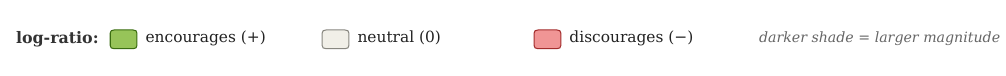}\\[-0.5ex]
    \includegraphics[width=0.82\linewidth,trim=0 8 100 38,clip]{figures/fig_privilege_flips_self_correction.pdf}
    \caption{\textbf{Privilege flips credit on self-correction cues, even
    when they lead to the correct answer.} Three windows from rollouts of
    a Qwen3-1.7B (Thinking) student trained against a Qwen3-8B (Thinking) teacher on
    OpenThoughts; the trajectory is identical under both teachers; only
    the teacher differs. Cells show the sign and magnitude of the
    per-token log-ratio
    $\log \pi_T(y_t \mid y_{<t}, x) - \log \pi_S(y_t \mid y_{<t}, x)$.
    \textbf{Top:} a rollout for \emph{``radius of a circle tangent to two
    concentric circles''} that ends at the wrong answer; privilege
    assigns negative advantage to \texttt{But wait} ($-1.96$, $-4.97$) and rewards the
    dismissive continuation \texttt{how can}. \textbf{Middle and bottom:}
    two windows from a rollout that reaches the \emph{correct} answer.
    Privilege still suppresses every self-correction cue, most extremely
    \texttt{but} ($-12.70$) and the hedge \texttt{Maybe} ($-1.64$). The
    privileged teacher's positive credit moves
    away from the moments where the student notices something is off and
    onto the locally fluent continuation.}
    \label{fig:privilege-flips}
\end{figure}

Figure~\ref{fig:privilege-flips} visualizes the reversal. The trajectory is held
fixed and scored by two teachers: an unprivileged teacher that sees the same
prompt as the student, and a privileged teacher that also sees answer
information. In a rollout that ends at the wrong answer, the privileged teacher
strongly suppresses the self-correction cue \texttt{But wait} ($-1.96$,
$-4.97$) and shifts positive credit toward the locally fluent continuation
\texttt{how can}. The same pattern appears in windows from a rollout that
reaches the correct answer. The privileged teacher suppresses self-correction
cues such as \texttt{but} ($-12.70$) and \texttt{Maybe} ($-1.64$), even though
the resulting trajectory reaches the right answer.

\begin{table*}[t]
\centering
\caption{\textbf{Gold-demonstration context suppresses epistemic-token usage more than vanilla OPD.}
This token-level companion to Table~\ref{tab:qwen17_openthoughts_opsd_opd} reports marker usage by \qwenonesevenb{} on OpenThoughts. The aggregate epistemic-token density is the fraction of generated tokens in the epistemic-marker set. The remaining columns report occurrences per 1,000 generated tokens for representative revision and search markers such as \textit{wait}, \textit{recall}, \textit{check}, and \textit{hmm}. Vanilla OPD leaves the aggregate density essentially unchanged, while the gold-demonstration variant lowers both aggregate density and several named markers.}
\label{tab:qwen17_openthoughts_epistemic_tokens}
\scriptsize
\setlength{\tabcolsep}{3pt}
\renewcommand{\arraystretch}{1.12}

\begin{tabular*}{\textwidth}{@{\extracolsep{\fill}}lcccccccc@{}}
\toprule
\textbf{Method}
& \shortstack{\textbf{Epistemic token}\\[-1pt]\textbf{density}}
& \textbf{wait}
& \textbf{recall}
& \textbf{okay}
& \textbf{altern}
& \textbf{check}
& \textbf{verify}
& \textbf{hmm} \\
\midrule
\textbf{Base}
& 1.080\%
& 3.85
& 1.11
& 2.17
& 0.64
& 1.33
& 0.25
& 1.45 \\
\textbf{+OPD}
& 1.074\%
& 3.79
& 1.04
& 2.17
& 0.72
& 1.19
& 0.16
& 1.68 \\
\textbf{+OPD gold demo}
& 0.850\%
& 2.54
& 0.96
& 2.07
& 0.66
& 1.05
& 0.10
& 1.11 \\
\bottomrule
\end{tabular*}
\end{table*}

Table~\ref{tab:qwen17_openthoughts_epistemic_tokens} reports the realized
density of the same marker set in \qwenonesevenb{} evaluation rollouts. Vanilla
OPD leaves the aggregate epistemic-token density nearly unchanged
($1.080\% \to 1.074\%$), while the gold-demonstration variant lowers it to $0.850\%$.
The reduction appears on several individual markers, including \textit{wait}
($3.85 \to 2.54$ occurrences per 1,000 generated tokens) and \textit{hmm}
($1.45 \to 1.11$).

The same pattern appears before sampling, in the probability mass assigned to
epistemic markers. Vanilla OPD leaves marginal marker mass essentially
unchanged, while the gold-demonstration teacher lowers it, with the largest
drops on \textit{wait}, \textit{recall}, \textit{altern}, and \textit{hmm}.
Full numbers and token-mask controls are in Appendix~\ref{app:opd-ablations}.

\subsection{The trained student produces fewer deliberation markers}
\label{sec:trained-student-deliberation}

The token-level analysis suggests that students trained with privileged context
should produce fewer deliberation markers at evaluation. We test this by
comparing paired base and \opsd{} rollouts with the same model, benchmark,
problem, and sample index (see Appendix~\ref{app:deliberation-markers}). The
analysis uses the five thinking models in the OpenThoughts 15k comparison on
AIME24, AIME25, and HMMT25, giving 7,200 paired rollouts.

\begin{table}[t]
\centering
\caption{\textbf{\opsd{} reduces explicit deliberation markers in paired thinking-model rollouts.}
We compare base and \opsd{} rollouts matched by model, benchmark, problem, and sample index across the five thinking models in the OpenThoughts 15k comparison. Counts are normalized per 1,000 generated tokens. Deltas are \opsd{} minus base, with confidence intervals from a clustered bootstrap over model--benchmark--problem clusters. All three marker families decrease after length normalization.}
\label{tab:deliberation_markers}
\small
\setlength{\tabcolsep}{4pt}
\renewcommand{\arraystretch}{1.12}
\begin{tabular*}{\linewidth}{@{\extracolsep{\fill}}lcccc@{}}
\toprule
\textbf{Marker family}
& \textbf{Base /1k}
& \textbf{\opsd{} /1k}
& \textbf{$\Delta$ /1k}
& \textbf{Raw $\Delta$} \\
\midrule
Verification
& 1.63
& 1.35
& $-0.28$ [$-0.32$, $-0.25$]
& $-8.8$ [$-9.5$, $-8.1$] \\
Backtracking
& 6.45
& 6.29
& $-0.16$ [$-0.32$, $-0.02$]
& $-23.5$ [$-29.0$, $-19.1$] \\
Hedging
& 3.79
& 3.62
& $-0.17$ [$-0.25$, $-0.10$]
& $-17.0$ [$-18.5$, $-15.5$] \\
\bottomrule
\end{tabular*}
\end{table}

Table~\ref{tab:deliberation_markers} counts three families of deliberation
markers: verification markers such as \textit{check} and \textit{verify},
backtracking markers such as \textit{wait}, \textit{actually}, and
\textit{wrong}, and hedging markers such as \textit{maybe} and
\textit{seems}. \opsd{} reduces the raw count of all three marker families, and
the reduction survives length normalization. Verification markers drop from
$1.63$ to $1.35$ per 1,000 generated tokens (95\% CI $[-0.32,-0.25]$), with
smaller but still negative shifts in backtracking and hedging. The length
collapse of Section~\ref{sec:results} is therefore accompanied by a
within-rollout reduction in deliberation markers, not just shorter rollouts
containing them at the same density.

Appendix~\ref{app:sd-zero-srt} gives complementary evidence from SD-Zero. The
self-revision stage alone helps the thinking model, but the subsequent \opsd{}
stage reverses that gain.

These diagnostics do not establish a causal explanation for the accuracy drop.
Rather, they identify a consistent behavioral pattern associated with the drop.
Privileged-context distillation degrades long-budget thinking behavior; the
degradation grows with the density of privileged teacher context; privileged
context lowers fork rates and raises lock rates in thinking-model rollouts;
fixed-trajectory scoring shows sign reversals on self-correction cues; and
trained students emit fewer deliberation markers. Together, these observations
support our interpretation of fork suppression as a failure mode of privileged
token-level supervision in thinking models, while leaving open whether it is the
primary cause of the observed accuracy degradation.

\section{Related Work}
\label{sec:relatedwork}

Self-distillation broadly refers to distillation settings where the teacher is
derived from the student model, often with the same architecture or checkpoint.
Our work focuses on recent privileged-context on-policy self-distillation
methods, where the self-teacher is additionally given information unavailable
to the student, such as answers, demonstrations, or feedback. Prior work has
reported gains from \opsd{}
\citep{zhao2026selfdistilledreasoneronpolicyselfdistillation}, SDFT
\citep{shenfeld2026self}, SDPO \citep{hubotter2026reinforcement}, SD-Zero
\citep{he2026selfdistillationzero}, and related recipes, especially for
instruction-tuned models, continual learning, or reasoning compression. We
show the complementary result that, in long-budget thinking models, privileged
token-level feedback can degrade the search behavior that enables test-time
reasoning. Concurrent work has linked self-distillation failures to
suppression of epistemic verbalization
\citep{kim2026doesselfdistillationsometimesdegrade}; our analysis studies the
broader fork-suppression mechanism using budget curves, fork/lock diagnostics,
fixed-trajectory token scoring, and trained-student marker shifts. We provide
an expanded discussion of related work in
Appendix~\ref{app:related-work-continued}.

\section{Discussion}
\label{sec:Discussion}

These findings suggest that the interaction between privileged supervision and
test-time search is more delicate for thinking models than for instruction-tuned
models. In our experiments, privileged-context OPD reduces the long-budget gains
of thinking models even when closely related unprivileged OPD improves the same
student. The token-level analyses provide a plausible explanation. Conditioning
the teacher on information unavailable to the student changes the learning
signal at forking positions, high-entropy decision points where different
continuations can send the rollout down different reasoning paths. Lexical
markers such as ``wait'' or ``maybe'' make some of these positions visible, but
the concern is broader than uncertainty language. Privileged feedback may
reshape which branches of the student's search are reinforced. Thus, the
failure mode is not simply that privileged distillation shortens responses or
suppresses epistemic markers, but that it can change the token-level structure
of the reasoning process. Future self-distillation methods for strong
thinking models may need to account for forking positions explicitly, so that
privileged teacher feedback improves solution-directed behavior without
collapsing branch-relevant test-time search.

More broadly, our results speak to \opsd{}-style training for long-horizon
agents. In our experiments, long-rollout degradation depended on the form of
privileged teacher-side context. Short final-answer-only context preserved
long-budget behavior better than a full gold demonstration. Reducing
privileged-context density therefore provides a partial mitigation. This
mitigation is relevant to emerging long-horizon
agent training recipes. For example, the
Composer 2.5 training stack uses targeted textual feedback, where a \emph{short}
hint is inserted into the teacher-side context and the original-context policy is
distilled toward the hinted distribution \citep{cursor2026composer25}.
Long-horizon \opsd{}-style training may therefore be viable, but likely requires
controlling both how much privileged information the teacher receives and where
the resulting token-level signal is applied.

\section{Limitations}
\label{ap:limitations}

Our analysis has several limitations. First, it does not fully isolate the mechanism or establish causality. Established work examines the importance of ``forking'' tokens in thinking-model reasoning, and we find that \opsd{} routinely suppresses them. Second, existing \opsd{} methods do not reliably improve thinking models, and we offer no solution. How to use privileged information for thinking-model distillation remains open. Third, compute limits preclude longer training completions matched to the 32k--38k evaluation budgets. Existing controls (unprivileged and privileged OPD, sparse and dense privileged context, and thinking-disabled and thinking-enabled training) nevertheless show that the short 4,096-token training cap is not by itself sufficient to explain the degradation. Fourth, we focus on verifiable math, where success and failure are measurable. We do not test continual learning, a proposed privileged-context self-distillation use case for non-thinking variants \citep{shenfeld2026self}.

\clearpage
\renewcommand{\acksection}{\section*{Acknowledgements}}
\begin{ack}
We thank Patrick Shafto for helpful discussions. SA acknowledges support from
DARPA through the AIQ and expMath programs, NSF, ONR, and the Schmidt
Foundation. The authors declare no competing interests.
\end{ack}

\bibliographystyle{plainnat}
\bibliography{references}

\clearpage
\appendix
\section{Expanded Related Work}
\label{app:related-work-continued}

\paragraph{Continual Learning.}
Model weights are updated during pre-training and post-training, but are then often deployed as static artifacts for months to serve traffic. This can quickly lead to a gap between in-weight knowledge and relevant real-world skills and information. \textit{Continual learning} aims to address this very gap \citep{de2021continual, kudithipudi2022biological, hadsell2020embracing, wang2024comprehensive}. Recently, self-distillation methods have claimed to allow for near seamless continual learning on new tasks while avoiding catastrophic forgetting common in other continual learning approaches \citep{shenfeld2026self}. This line of work claims that the policy being updated is minimally changed when trained using on-policy distillation from a teacher shown privileged information.

\paragraph{On-Policy Distillation.}
Distillation methods aim to impart richer knowledge into a student model
(usually from an already trained teacher) compared to standard supervised
learning methods \citep{hinton2015distilling, gou2021knowledge,
agarwal2024policy}. For language modeling tasks, several variants of
distillation exist. \textit{Sequence distillation} uses natural language outputs
of one model to directly fine-tune a student model. \textit{Knowledge
distillation} goes one step further and trains a student using the
probabilities that a teacher assigns to its generated sequence. \textit{On-policy
distillation} has emerged as an alternative to standard knowledge distillation
wherein the sequence being scored, and subsequently used for training, is
generated \textit{on-policy} by the student model. This approach alleviates
observed issues between train-test distribution shift that can arise in standard
knowledge distillation.

\paragraph{Self-Improving Distillation.}
Generally speaking, knowledge distillation methods, including sequence
distillation and on/off-policy distillation, require a stronger teacher in order
to improve the student's performance. This can induce a heavy computational
overhead for researchers and also raises an important question: can a model be
used to improve itself without the need for a stronger teacher? Several recent
approaches have attempted to answer this question in the affirmative.
On-Policy Self-Distillation (OPSD)
\citep{zhao2026selfdistilledreasoneronpolicyselfdistillation} utilizes a single
model architecture to act as both teacher and student: the teacher policy is
conditioned on privileged information, such as a ground-truth answer to a math
problem, and is used to score on-policy generations of a student without the
privileged information. It is worth noting that recent updates to the OPSD
method specifically disable per-token divergence feedback on deliberation
tokens, which is necessary to stabilize training. Self-Distillation Fine-Tuning
(SDFT) \citep{shenfeld2026self} proposes a similar method but focuses on
integrating new knowledge corpora for continual learning rather than verifiable
math questions.

We focus on self-distillation methods in the previous vein: methods that use
ground truth privileged information to directly condition a teacher's
token-level feedback to train a student. However, other flavors of
self-distillation methods may target different objectives and mechanisms. For
example, CRISP \citep{sang2026policy} focuses on reducing thinking trace lengths
by distilling a teacher's concise reasoning into a student model.
Self-distillation has also been attempted where the teacher, instead of being
given privileged information or different steering prompts, is instead sampled
at different temperatures \citep{zhang2026embarrassingly}. Approaches like
Self-Distillation Policy Optimization (SDPO)
\citep{hubotter2026reinforcement} also use privileged information but augment
the teacher policy with environmental feedback like error messages instead of
gold, ground-truth information like OPSD or SDFT.

\paragraph{Self-Distillation Failure Modes.}
Concurrent work \citep{kim2026doesselfdistillationsometimesdegrade} studies why
self-distillation can sometimes degrade mathematical reasoning, and attributes
the degradation to suppression of epistemic verbalization under richer teacher
conditioning. Their analysis emphasizes context richness and task coverage:
richer conditioning produces shorter, more confident traces with fewer
uncertainty expressions, which can help narrow in-domain settings but hurt OOD
math generalization. Viewed through our framework, epistemic-verbalization
suppression is a visible lexical subset of fork suppression: overt uncertainty
markers expose some high-entropy fork positions, but many branch-relevant
decisions occur at ordinary mathematical, connective, or formatting
continuations. Our work is therefore complementary. We focus on long-rollout
thinking models and ask how privileged token-level feedback changes the
high-entropy decision points that support test-time search. Rather than treating
epistemic markers as the primary object of study, we analyze fork-like positions
in the teacher distribution, token-level signal reversal along fixed student
trajectories, and the resulting loss of long-budget test-time compute gains.

\paragraph{Forking and Exploration in Reasoning Traces.}
Reasoning ability from a token-level perspective has been analyzed in several
works that have found that some tokens, called forking tokens, can have an
outsized influence on the downstream success of a reasoning trace
\citep{bigelow2024forking, lin2024critical, vassoyan2025ignore,
zhang2026embarrassingly}. These critical tokens often occur at high-entropy
positions in a model's generation and control where and how the reasoning
branches into different paths and alternative strategies. Our work
contextualizes failures of self-improvement distillation methods in the setting
of forking suppression.

\paragraph{Privileged Feedback in RL.}
Standard RL setups may face exploration bottlenecks on complex reasoning tasks where correct rollouts can be rare. To overcome this, several methods in RL also aim to incorporate privileged information into the exploration stage. For example, Privileged On-Policy Exploration (POPE) \citep{qu2026pope} and PrefixRL \citep{setlur2026reuse} utilize some privileged information to guide on-policy exploration. Hybrid approaches like HDPO \citep{ding2026hdpo} augment standard RL training with targeted privileged self-distillation in certain cases. Similar frameworks like Self-Distilled RLVR (RLSD) \citep{yang2026self} incorporate self-distillation to guide update magnitudes of standard RLVR training. Our work sets aside RL training mechanisms to focus on the token-level dynamics of privileged distillation---a harm that hybrid approaches may be able to mitigate.

\clearpage
\section{Experimental Details}
\label{app:experimental-details}

This appendix records the data, training, and evaluation settings used for the
experiments in the main text. Unless a table or paragraph states otherwise, all
reported \opsd{} runs use the defaults in Table~\ref{tab:opsd-training-hparams}
and all evaluations use the generation and grading protocol in
Table~\ref{tab:eval-hparams}.

\subsection{Data}
\label{app:data}

\paragraph{OpenThoughts math.}
We use a cleaned version of \texttt{Ashkchamp/Openthoughts\_math\_filtered\_30K}.
The cleaning script removes system turns, remaps thought delimiters to
\texttt{<think>} and \texttt{</think>}, removes explicit solution delimiters,
and appends the instruction to return the final answer in \texttt{\textbackslash
boxed\{\}}. We use a 15K subset of the data for training. The privileged teacher context is taken from the \texttt{solution} column.

\paragraph{Countdown.}
For Countdown, we use \texttt{jasonrqh/Countdown-CoT-20k}. We select a 15K subset for training and reserve an additional
500 examples as a held-out evaluation set.

\begin{table}[!htbp]
    \centering
    \small
    \caption{Training and evaluation data used in the experiments.}
    \label{tab:data-summary}
    \begin{tabular}{ll>{\raggedright\arraybackslash}p{0.58\linewidth}}
        \toprule
        Data & Rows & Use and fields \\
        \midrule
        OpenThoughts cleaned & 29,439 &
        Source pool after cleaning; columns include \texttt{problem},
        \texttt{solution}, and \texttt{Answer}. \\
        OpenThoughts 15k & 15,000 &
        Main OpenThoughts \opsd{} training subset; prompt =
        \texttt{problem}, privileged context = \texttt{solution}. \\
        Countdown train & 15,000 &
        Countdown \opsd{} training; prompt =
        \texttt{datapoint\_input\_text}, privileged context =
        \texttt{response\_suffix}. \\
        Countdown eval & 500 &
        Held-out Countdown evaluation. \\
        AIME24/AIME25/HMMT25 & 30 each &
        Math competition evaluations. \\
        \bottomrule
    \end{tabular}
\end{table}

\subsection{Existing Asset Licenses}
\label{app:asset-licenses}

Table~\ref{tab:asset-licenses} summarizes the existing datasets, model
checkpoints, and software assets used in our experiments. We use these assets
for training, evaluation, or implementation, and do not redistribute
third-party datasets or model checkpoints.

\begin{table}[!htbp]
    \centering
    \scriptsize
    \caption{Existing assets used in this work.}
    \label{tab:asset-licenses}
    \setlength{\tabcolsep}{3pt}
    \renewcommand{\arraystretch}{1.15}
    \begin{tabular}{>{\raggedright\arraybackslash}p{0.23\linewidth}
                    >{\raggedright\arraybackslash}p{0.28\linewidth}
                    >{\raggedright\arraybackslash}p{0.18\linewidth}
                    >{\raggedright\arraybackslash}p{0.23\linewidth}}
        \toprule
        Asset & Source & License / terms & Use \\
        \midrule
        \texttt{Ashkchamp/Openthoughts\_math\_filtered\_30K} &
        \url{https://huggingface.co/datasets/Ashkchamp/Openthoughts_math_filtered_30K} &
        No explicit license metadata listed on the dataset page at time of access &
        Source pool for the OpenThoughts math training subset; not redistributed. \\
        \texttt{open-thoughts/OpenThoughts-114k} &
        \url{https://huggingface.co/datasets/open-thoughts/OpenThoughts-114k} &
        Apache-2.0 &
        Upstream OpenThoughts dataset related to the math training data; not redistributed. \\
        \texttt{jasonrqh/Countdown-CoT-20k} &
        \url{https://huggingface.co/datasets/jasonrqh/Countdown-CoT-20k} &
        MIT &
        Countdown training and held-out in-domain evaluation data. \\
        Qwen3 checkpoints &
        \url{https://huggingface.co/Qwen} &
        Apache-2.0 &
        Base instruction and thinking models used for training and evaluation. \\
        OLMo-3 checkpoints &
        \url{https://huggingface.co/allenai/Olmo-3-7B-Think} &
        Apache-2.0; Ai2 Responsible Use Guidelines &
        Base instruction and thinking models used for training and evaluation. \\
        \texttt{idanshen/Self-Distillation} &
        \url{https://github.com/idanshen/Self-Distillation} &
        No explicit software license file found at time of access &
        Reference/adapted implementation code for self-distillation; not redistributed. \\
        \texttt{vLLM} &
        \url{https://github.com/vllm-project/vllm} &
        Apache-2.0 &
        Inference engine used for generation and evaluation. \\
        AIME and HMMT problems &
        Public competition materials &
        Public competition materials; not redistributed &
        Out-of-domain math evaluation benchmarks. \\
        \bottomrule
    \end{tabular}
\end{table}

\subsection{Prompt Templates and Privileged Context Examples}
\label{app:prompt-templates}

\newcommand{\promptvar}[1]{\texttt{\textcolor{RoyalBlue!70!black}{\{#1\}}}}
\newcommand{\promptcard}[3]{%
    \begin{center}
    \setlength{\fboxsep}{7pt}%
    \fcolorbox{#1!55!black}{#1!5}{%
        \begin{minipage}{0.92\linewidth}
        \raggedright
        \textbf{\textcolor{#1!55!black}{#2}}\par\vspace{4pt}
        \small #3
        \end{minipage}%
    }%
    \end{center}
}

We fill the following templates before applying each model's chat template.
For standard \opsd{} training, the student receives only the problem.
\promptcard{RoyalBlue}{Student prompt template}{%
{\ttfamily
\promptvar{problem}
}
}

The privileged teacher receives the same problem plus an example response.
\promptcard{OliveGreen}{Privileged-teacher prompt template}{%
{\ttfamily
\promptvar{problem}\par\medskip
This is an example for a response to the question:\par\medskip
\promptvar{Answer}\par\medskip
Now answer with a response of your own, including the thinking process.
}
}
Here \promptvar{Answer} denotes the privileged response used as teacher
context. In dense gold-demonstration runs, this field contains a full reference
solution; in sparse final-answer-only runs, it contains only the boxed final
answer.

For the conciseness-control experiment in
Figure~\ref{fig:openthoughts_qwen8b_concise_budget}, the teacher is prompted
without gold context.
\promptcard{BurntOrange}{Conciseness-control teacher prompt}{%
{\ttfamily
Solve the following math problem concisely and correctly. Be direct -- avoid
unnecessary elaboration, redundant steps, or restating the problem. Focus only
on the key reasoning steps needed to reach the answer.\par\medskip
\promptvar{problem}
}
}

\clearpage
\paragraph{Sparse and dense context example.}
The dense and sparse conditions use the same privileged-teacher wrapper above;
they differ only in what is inserted into the \promptvar{Answer} slot. The
following filled examples use the same problem.

\promptcard{OliveGreen}{Dense teacher prompt, filled}{%
{\ttfamily
Given real numbers \(a, b, c\) and a positive number \(\lambda\) such that the
polynomial \(f(x) = x^3 + a x^2 + b x + c\) has three real roots \(x_1, x_2,
x_3\), and the conditions \(x_2 - x_1 = \lambda\) and
\(x_3 > \frac{1}{2}(x_1 + x_2)\) are satisfied, find the maximum value of
\[
\frac{2 a^3 + 27 c - 9 a b}{\lambda^3}.
\]
\par\medskip
This is an example for a response to the question:\par\medskip
We begin by analyzing the function \(f(x) = x^3 + a x^2 + b x + c\), which has
three real roots \(x_1, x_2, x_3\). We are given the following conditions:
\(x_2 - x_1 = \lambda\) and
\(x_3 > \frac{1}{2}(x_1 + x_2)\). We aim to find the maximum value of
\[
\frac{2a^3 + 27c - 9ab}{\lambda^3}.
\]
\par\medskip
Transform the polynomial to remove the quadratic term.
Substitute \(x = y - \frac{a}{3}\) into \(f(x)\):
\[
\begin{array}{rcl}
F(y) & = & f\left(y - \frac{a}{3}\right) \\
     & = & \left(y - \frac{a}{3}\right)^3
           + a \left(y - \frac{a}{3}\right)^2
           + b \left(y - \frac{a}{3}\right) + c \\
     & = & y^3 - \left(\frac{a^2}{3} - b\right)y
           + \frac{1}{27}(2a^3 + 27c - 9ab).
\end{array}
\]
\par\medskip
Identify the new roots of \(F(y)\).
Let the roots of \(F(y)\) be \(y_1, y_2, y_3\). We know
\(y_i = x_i + \frac{a}{3}\). Using Vieta's formulas,
\[
y_1 + y_2 + y_3 = 0,\qquad
y_1 y_2 y_3 = -\frac{1}{27}(2a^3 + 27c - 9ab).
\]
\par\medskip
\textit{[Middle of the gold demonstration omitted for clarity of the example.]}
\par\medskip
Conclusion:
\[
\boxed{\frac{3\sqrt{3}}{2}}
\]
\par\medskip
Now answer with a response of your own, including the thinking process.
}
}

\promptcard{OliveGreen}{Sparse final-answer-only teacher prompt, filled}{%
{\ttfamily
Given real numbers \(a, b, c\) and a positive number \(\lambda\) such that the
polynomial \(f(x) = x^3 + a x^2 + b x + c\) has three real roots \(x_1, x_2,
x_3\), and the conditions \(x_2 - x_1 = \lambda\) and
\(x_3 > \frac{1}{2}(x_1 + x_2)\) are satisfied, find the maximum value of
\[
\frac{2 a^3 + 27 c - 9 a b}{\lambda^3}.
\]
\par\medskip
This is an example for a response to the question:\par
\[
\boxed{\frac{3\sqrt{3}}{2}}
\]
\par\medskip
Now answer with a response of your own, including the thinking process.
}
}

\subsection{OPSD Training}
\label{app:opsd-training}

For each training example, the student is prompted with the task input alone.
The teacher is initialized from the same base checkpoint, but receives the same
task input plus privileged supervision. The trainer samples completions
on-policy and minimizes token-level divergence between teacher and student
distributions on those sampled completion tokens.

\begin{table}[!htbp]
    \centering
    \small
    \caption{Default \opsd{} training hyperparameters.}
    \label{tab:opsd-training-hparams}
    \begin{tabular}{ll}
        \toprule
        Hyperparameter & Value \\
        \midrule
        Objective & Generalized KL/JSD distillation on sampled completion tokens \\
        Distillation mixture $\alpha$ & 0.5 (JSD) \\
        Teacher synchronization & Disabled for reported regular runs \\
        Completions per prompt & 1 \\
        Training sampling temperature / top-$p$  & 1.0 / 1.0  \\
        Epochs & 1 \\
        Effective batch size & 64 \\
        Per-device batch size & 1 \\
        Gradient accumulation & 8 \\
        Optimizer learning-rate schedule & Cosine decay \\
        Warmup ratio & 0.1 \\
        Max gradient norm & 1.0 \\
        Weight decay & 0.0 \\
        Precision & bf16 \\
        Distributed training & FSDP full-shard auto-wrap \\
        Gradient checkpointing & Enabled \\
        Seed & 42 \\
        Max prompt length & 25,000 tokens \\
        Max completion length & 4,096 tokens unless noted otherwise \\
        \bottomrule
    \end{tabular}
\end{table}

\begin{table}[!htbp]
    \centering
    \small
    \caption{Model-specific \opsd{} training settings and deviations from
    Table~\ref{tab:opsd-training-hparams}.}
    \label{tab:model-training-hparams}
    \resizebox{\textwidth}{!}{%
    \begin{tabular}{llllll}
        \toprule
        Model(s) & Training data & LR & Max completion & vLLM train gen. & Hardware \\
        \midrule
        Qwen3-1.7B, Qwen3-4B, Qwen3-4B-Thinking-2507,
        Qwen3-4B-Instruct-2507 &
        OpenThoughts 15k & $5{\times}10^{-6}$ & 4,096 & Yes &
        8 H100, 80GB/GPU \\
        Qwen3-8B &
        OpenThoughts 15k & $2{\times}10^{-6}$ & 4,096 & Yes &
        8 H100, 80GB/GPU \\
        OLMo-3-7B-Instruct, OLMo-3-7B-Think &
        OpenThoughts 15k & $5{\times}10^{-6}$ & 4,096 & Yes &
        8 H100, 80GB/GPU \\
        Qwen3-4B-Thinking-2507, Qwen3-4B-Instruct-2507,
        OLMo-3-7B-Instruct, OLMo-3-7B-Think &
        Countdown 15k & $5{\times}10^{-6}$ & 4,096 & Yes &
        8 H100, 80GB/GPU \\
        \bottomrule
    \end{tabular}%
    }
\end{table}

\subsection{Evaluation}
\label{app:evaluation}

For AIME24, AIME25, and HMMT25, we generate 16 samples per problem on 30
problems per benchmark. Generation is sharded over 8 jobs and uses a maximum
generation length of 38,912 tokens. The merged generation file therefore
contains 480 rows for each AIME/HMMT benchmark. Countdown uses the same
16-sample evaluation protocol on the 500-example held-out split when reported.

\begin{table}[!htbp]
    \centering
    \small
    \caption{Evaluation generation hyperparameters.}
    \label{tab:eval-hparams}
    \begin{tabular}{lll}
        \toprule
        Model/eval group & Temperature & Top-$p$ \\
        \midrule
        Qwen3 thinking-style models & 0.6 & 0.95 \\
        Qwen3-4B-Instruct-2507 & 0.7 & 0.8 \\
        OLMo-3-7B-Instruct/Think & 0.6 & 0.95 \\
        \bottomrule
    \end{tabular}
\end{table}

\begin{table}[!htbp]
    \centering
    \small
    \caption{Shared evaluation settings.}
    \label{tab:eval-shared-hparams}
    \begin{tabular}{ll}
        \toprule
        Setting & Value \\
        \midrule
        Samples per problem & 16 \\
        Shards & 8 \\
        Maximum generation length & 38,912 tokens \\
        Eval engine & vLLM \\
        Eval precision & bf16 \\
        vLLM GPU memory utilization & 0.9 \\
        Eval top-$k$ & -1 \\
        Prompt format & Model chat template with generation prompt \\
        Metric file & \texttt{metrics.json} written next to merged JSONL \\
        \bottomrule
    \end{tabular}
\end{table}

\paragraph{Metrics.}
The reported avg@16 \texttt{accuracy} is the mean correctness over generated
samples: for each problem, we average correctness across its 16 sampled rollouts,
then average across problems. This is equivalent to empirical single-sample
correctness under the evaluation sampling distribution, but is distinct from
unbiased pass@16.
For $k > 1$, pass@$k$ is computed
with the unbiased estimator. For each problem with $n$ samples and $c$ correct
samples, the contribution is 1 if $n-c < k$ and otherwise
\[
1 - \prod_{i=0}^{k-1} \frac{n-c-i}{n-i}.
\]
We report pass@$k$ only for $k \leq n$.

\subsection{Deliberation Marker Analysis}
\label{app:deliberation-markers}

To test whether \opsd{} changes explicit deliberation language in model
reasoning, we compare paired base and \opsd{} rollouts for the same model,
benchmark, problem, and sample (see Table~\ref{tab:deliberation_markers_full}). The analysis uses the five thinking models
in the OpenThoughts 15k comparison on AIME24, AIME25, and HMMT25. With 16
samples for each of 30 problems on each benchmark, this gives 7,200 paired
rollouts.

For each response, we count occurrences of a fixed, case-insensitive lexicon of
deliberation markers. The marker families are verification markers, with
examples such as \textit{check}, \textit{verify}, \textit{double check}, and
\textit{make sure}; backtracking markers, with examples such as \textit{wait},
\textit{actually}, \textit{mistake}, \textit{wrong}, and \textit{another way};
and hedging markers, with examples such as \textit{maybe}, \textit{probably},
\textit{might}, and \textit{seems}. We report raw marker counts and marker
counts per 1,000 response tokens. Response-token counts are computed with the
cached model tokenizer.

\begin{table*}[t]
\centering
\caption{\textbf{Full deliberation-marker counts for paired base and \opsd{} rollouts.}
Raw columns report average marker counts per response. Normalized columns report occurrences per 1,000 generated tokens, using response-token counts computed with the cached model tokenizer. Deltas are \opsd{} minus base, with confidence intervals computed by clustered bootstrap over model--benchmark--problem clusters.}
\label{tab:deliberation_markers_full}
\small
\setlength{\tabcolsep}{4pt}
\renewcommand{\arraystretch}{1.12}
\begin{tabular*}{\textwidth}{@{\extracolsep{\fill}}lcccccc@{}}
\toprule
\textbf{Marker family}
& \textbf{Base raw}
& \textbf{\opsd{} raw}
& \textbf{Raw $\Delta$}
& \textbf{Base /1k}
& \textbf{\opsd{} /1k}
& \textbf{$\Delta$ /1k} \\
\midrule
Verification
& 26.1
& 17.3
& $-8.8$ [$-9.5$, $-8.1$]
& 1.63
& 1.35
& $-0.28$ [$-0.32$, $-0.25$] \\
Backtracking
& 124.2
& 100.6
& $-23.5$ [$-29.0$, $-19.1$]
& 6.45
& 6.29
& $-0.16$ [$-0.32$, $-0.02$] \\
Hedging
& 72.2
& 55.2
& $-17.0$ [$-18.5$, $-15.5$]
& 3.79
& 3.62
& $-0.17$ [$-0.25$, $-0.10$] \\
\bottomrule
\end{tabular*}
\end{table*}

Deltas are \opsd{} minus base. Confidence intervals are computed by clustered
bootstrap over model--benchmark--problem clusters, so the 16 samples from the
same problem are not treated as fully independent. The mean response length in
this analysis is 19,395 tokens for base rollouts and 15,391 tokens after
\opsd{}. Since the counts are lexical proxies, we interpret them as evidence
about explicit deliberation markers in the generated traces, not as direct
measurements of latent uncertainty or confidence.

\clearpage
\section{Accuracy Confidence Intervals and Full Pass@k Results}
\label{app:full-passk-results}

\subsection{Countdown Think-vs-Instruct Results}

\begin{table*}[t]
\centering
\caption{\textbf{\opsd{} helps instruct models more reliably than thinking models on Countdown.}
We use \opsd{} to train various models on Countdown and evaluate on held-out Countdown data, AIME24, AIME25, and HMMT25. Entries report avg@16 accuracy, and the Average column averages the four benchmarks. Instruct Qwen and OLMo models improve under \opsd{}, while the corresponding thinking models show smaller or mixed gains.}
\label{tab:countdown_think_vs_instruct}
\small
\setlength{\tabcolsep}{5pt}
\renewcommand{\arraystretch}{1.15}

\newcommand{\metrichead}[1]{%
  \textbf{#1}%
}

\begin{tabular}{@{}llccccc@{}}
\toprule
\textbf{Model} &
& \metrichead{Countdown}
& \metrichead{AIME24}
& \metrichead{AIME25}
& \metrichead{HMMT25}
& \metrichead{Average} \\
\midrule

\multirow{2}{*}{\textbf{\qwenfourbinstruct{}}}
& \textbf{Base}
& 0.841
& \textbf{0.604}
& 0.463
& \textbf{0.304}
& 0.553 \\
&
\textbf{+\opsd{}}
& \textbf{0.865}
& 0.594
& \textbf{0.483}
& 0.294
& \textbf{0.559} \\
\midrule

\multirow{2}{*}{\textbf{\qwenfourbthink{}}}
& \textbf{Base}
& 0.945
& \textbf{0.804}
& \textbf{0.804}
& \textbf{0.552}
& \textbf{0.776} \\
&
\textbf{+\opsd{}}
& \textbf{0.947}
& 0.800
& 0.775
& 0.537
& 0.765 \\
\midrule

\multirow{2}{*}{\textbf{\olmosevenbinstruct{}}}
& \textbf{Base}
& 0.775
& \textbf{0.525}
& \textbf{0.415}
& 0.237
& 0.488 \\
&
\textbf{+\opsd{}}
& \textbf{0.814}
& 0.510
& 0.394
& \textbf{0.256}
& \textbf{0.494} \\
\midrule

\multirow{2}{*}{\textbf{\olmosevenbthink{}}}
& \textbf{Base}
& 0.879
& 0.719
& 0.667
& \textbf{0.452}
& 0.679 \\
&
\textbf{+\opsd{}}
& \textbf{0.890}
& \textbf{0.742}
& \textbf{0.698}
& \textbf{0.452}
& \textbf{0.695} \\

\bottomrule
\end{tabular}
\end{table*}

\subsection{Paired Bootstrap Confidence Intervals}
\label{app:bootstrap-cis}

Tables~\ref{tab:countdown_bootstrap_cis_full},
\ref{tab:openthoughts_bootstrap_cis_full}, and
\ref{tab:opd_bootstrap_cis_full} report paired 95\% bootstrap confidence
intervals for the avg@16 deltas in
Tables~\ref{tab:countdown_think_vs_instruct},
\ref{tab:openthoughts_thinking_models}, and
\ref{tab:qwen17_openthoughts_opsd_opd}, respectively. For this metric, we first
compute each problem's contribution as the mean correctness over its 16 sampled
rollouts. For each bootstrap replicate, we
resample evaluation problems with replacement within each benchmark, preserving
the paired measurements for the two methods being compared. We then compute the
method delta within each benchmark and average benchmark deltas using the same
aggregation as the corresponding table. Intervals are percentile intervals from
10,000 bootstrap replicates. These intervals reflect evaluation-problem
uncertainty for the observed 16-sample estimates, but not training-seed
variability. Thus, for a 30-problem benchmark, the bootstrap operates over 30
paired problem-level observations, not 480 rollout-level observations.

\begin{table*}[!htbp]
\centering
\caption{\textbf{\opsd{} helps instruct models more reliably than thinking models on Countdown.}
Companion to Table~\ref{tab:countdown_think_vs_instruct}; $\Delta$ rows report
paired bootstrap 95\% confidence intervals for avg@16 deltas.}
\label{tab:countdown_bootstrap_cis_full}
\scriptsize
\setlength{\tabcolsep}{3pt}
\renewcommand{\arraystretch}{1.12}
\providecommand{\cicell}[2]{\shortstack{#1\\[-1pt]{\scriptsize #2}}}
\providecommand{\deltalabel}[1]{\shortstack[l]{\textbf{$\Delta$ (#1)}\\[-1pt]{\scriptsize $\Delta$ [95\% CI]}}}

\begin{tabular*}{\textwidth}{@{\extracolsep{\fill}}llccccc@{}}
\toprule
\textbf{Model} &
& \textbf{Countdown}
& \textbf{AIME24}
& \textbf{AIME25}
& \textbf{HMMT25}
& \textbf{Average} \\
\midrule

\multirow{3}{*}{\textbf{\qwenfourbinstruct{}}}
& \textbf{Base}
& 0.841 & 0.604 & 0.463 & 0.304 & 0.553 \\
& \textbf{+\opsd{}}
& 0.865 & 0.594 & 0.483 & 0.294 & 0.559 \\
& \deltalabel{\opsd{} - Base}
& \cicell{$+0.024$}{$[0.016,0.032]$}
& \cicell{$-0.010$}{$[-0.052,0.033]$}
& \cicell{$+0.021$}{$[-0.010,0.054]$}
& \cicell{$-0.010$}{$[-0.044,0.019]$}
& \cicell{$+0.006$}{$[-0.009,0.022]$} \\
\midrule

\multirow{3}{*}{\textbf{\qwenfourbthink{}}}
& \textbf{Base}
& 0.945 & 0.804 & 0.804 & 0.552 & 0.776 \\
& \textbf{+\opsd{}}
& 0.947 & 0.800 & 0.775 & 0.537 & 0.765 \\
& \deltalabel{\opsd{} - Base}
& \cicell{$+0.002$}{$[-0.010,0.014]$}
& \cicell{$-0.004$}{$[-0.027,0.019]$}
& \cicell{$-0.029$}{$[-0.058,-0.002]$}
& \cicell{$-0.015$}{$[-0.062,0.021]$}
& \cicell{$-0.011$}{$[-0.026,0.002]$} \\
\midrule

\multirow{3}{*}{\textbf{\olmosevenbinstruct{}}}
& \textbf{Base}
& 0.775 & 0.525 & 0.415 & 0.237 & 0.488 \\
& \textbf{+\opsd{}}
& 0.814 & 0.510 & 0.394 & 0.256 & 0.494 \\
& \deltalabel{\opsd{} - Base}
& \cicell{$+0.039$}{$[0.028,0.050]$}
& \cicell{$-0.015$}{$[-0.056,0.025]$}
& \cicell{$-0.021$}{$[-0.058,0.013]$}
& \cicell{$+0.019$}{$[-0.002,0.040]$}
& \cicell{$+0.006$}{$[-0.009,0.020]$} \\
\midrule

\multirow{3}{*}{\textbf{\olmosevenbthink{}}}
& \textbf{Base}
& 0.879 & 0.719 & 0.667 & 0.452 & 0.679 \\
& \textbf{+\opsd{}}
& 0.890 & 0.742 & 0.698 & 0.452 & 0.695 \\
& \deltalabel{\opsd{} - Base}
& \cicell{$+0.010$}{$[0.003,0.018]$}
& \cicell{$+0.023$}{$[-0.006,0.052]$}
& \cicell{$+0.031$}{$[0.002,0.067]$}
& \cicell{$0.000$}{$[-0.025,0.031]$}
& \cicell{$+0.016$}{$[0.004,0.030]$} \\

\bottomrule
\end{tabular*}
\end{table*}

\begin{table*}[!htbp]
\centering
\caption{\textbf{\opsd{} degrades thinking models across model families.}
Companion to Table~\ref{tab:openthoughts_thinking_models}; $\Delta$ rows report
paired bootstrap 95\% confidence intervals for avg@16 deltas.}
\label{tab:openthoughts_bootstrap_cis_full}
\scriptsize
\setlength{\tabcolsep}{3pt}
\renewcommand{\arraystretch}{1.12}
\providecommand{\cicell}[2]{\shortstack{#1\\[-1pt]{\scriptsize #2}}}
\providecommand{\deltalabel}[1]{\shortstack[l]{\textbf{$\Delta$ (#1)}\\[-1pt]{\scriptsize $\Delta$ [95\% CI]}}}

\begin{tabular*}{\textwidth}{@{\extracolsep{\fill}}llcccc@{}}
\toprule
\textbf{Model}
&
& \textbf{AIME24}
& \textbf{AIME25}
& \textbf{HMMT25}
& \textbf{Average} \\
\midrule

\multirow{3}{*}{\textbf{\qwenonesevenb{} (Thinking)}}
& \textbf{Base}
& 0.502 & 0.398 & 0.215 & 0.372 \\
& \textbf{+\opsd{}}
& 0.435 & 0.302 & 0.185 & 0.308 \\
& \deltalabel{\opsd{} - Base}
& \cicell{$-0.067$}{$[-0.125,-0.017]$}
& \cicell{$-0.096$}{$[-0.160,-0.042]$}
& \cicell{$-0.029$}{$[-0.062,0.002]$}
& \cicell{$-0.064$}{$[-0.095,-0.037]$} \\
\midrule

\multirow{3}{*}{\textbf{\qwenfourb{} (Thinking)}}
& \textbf{Base}
& 0.727 & 0.635 & 0.410 & 0.591 \\
& \textbf{+\opsd{}}
& 0.683 & 0.556 & 0.362 & 0.534 \\
& \deltalabel{\opsd{} - Base}
& \cicell{$-0.044$}{$[-0.100,0.004]$}
& \cicell{$-0.079$}{$[-0.138,-0.019]$}
& \cicell{$-0.048$}{$[-0.098,-0.002]$}
& \cicell{$-0.057$}{$[-0.088,-0.027]$} \\
\midrule

\multirow{3}{*}{\textbf{\qweneightb{} (Thinking)}}
& \textbf{Base}
& 0.758 & 0.700 & 0.448 & 0.635 \\
& \textbf{+\opsd{}}
& 0.721 & 0.613 & 0.400 & 0.578 \\
& \deltalabel{\opsd{} - Base}
& \cicell{$-0.037$}{$[-0.085,0.006]$}
& \cicell{$-0.087$}{$[-0.160,-0.027]$}
& \cicell{$-0.048$}{$[-0.098,0.000]$}
& \cicell{$-0.058$}{$[-0.090,-0.028]$} \\
\midrule

\multirow{3}{*}{\textbf{\qwenfourbthink{}}}
& \textbf{Base}
& 0.804 & 0.804 & 0.552 & 0.720 \\
& \textbf{+\opsd{}}
& 0.787 & 0.731 & 0.529 & 0.683 \\
& \deltalabel{\opsd{} - Base}
& \cicell{$-0.017$}{$[-0.052,0.017]$}
& \cicell{$-0.073$}{$[-0.121,-0.029]$}
& \cicell{$-0.023$}{$[-0.081,0.037]$}
& \cicell{$-0.037$}{$[-0.065,-0.010]$} \\
\midrule

\multirow{3}{*}{\textbf{\olmosevenbthink{}}}
& \textbf{Base}
& 0.719 & 0.667 & 0.452 & 0.612 \\
& \textbf{+\opsd{}}
& 0.715 & 0.652 & 0.446 & 0.604 \\
& \deltalabel{\opsd{} - Base}
& \cicell{$-0.004$}{$[-0.033,0.023]$}
& \cicell{$-0.015$}{$[-0.058,0.025]$}
& \cicell{$-0.006$}{$[-0.037,0.027]$}
& \cicell{$-0.008$}{$[-0.028,0.011]$} \\

\bottomrule
\end{tabular*}
\end{table*}

\begin{table*}[!htbp]
\centering
\caption{\textbf{Privileged teacher context, not on-policy distillation itself, degrades thinking-model performance.}
Companion to Table~\ref{tab:qwen17_openthoughts_opsd_opd}; $\Delta$ rows report
paired bootstrap 95\% confidence intervals for avg@16 deltas.}
\label{tab:opd_bootstrap_cis_full}
\scriptsize
\setlength{\tabcolsep}{3pt}
\renewcommand{\arraystretch}{1.12}
\providecommand{\cicell}[2]{\shortstack{#1\\[-1pt]{\scriptsize #2}}}
\providecommand{\deltalabel}[1]{\shortstack[l]{\textbf{$\Delta$ (#1)}\\[-1pt]{\scriptsize $\Delta$ [95\% CI]}}}

\begin{tabular*}{\textwidth}{@{\extracolsep{\fill}}llcccc@{}}
\toprule
\textbf{Model}
&
& \textbf{AIME24}
& \textbf{AIME25}
& \textbf{HMMT25}
& \textbf{Average} \\
\midrule

\multirow{7}{*}{\textbf{\qwenonesevenb{}}}
& \textbf{Base}
& 0.502 & 0.398 & 0.215 & 0.372 \\
& \textbf{+\opsd{}}
& 0.435 & 0.302 & 0.185 & 0.308 \\
& \textbf{+OPD}
& 0.540 & 0.385 & 0.252 & 0.392 \\
& \textbf{+OPD gold demo}
& 0.467 & 0.344 & 0.240 & 0.350 \\
& \deltalabel{OPD - Base}
& \cicell{$+0.037$}{$[-0.006,0.081]$}
& \cicell{$-0.013$}{$[-0.056,0.027]$}
& \cicell{$+0.037$}{$[0.013,0.065]$}
& \cicell{$+0.021$}{$[-0.001,0.042]$} \\
& \deltalabel{OPD gold demo - OPD}
& \cicell{$-0.073$}{$[-0.123,-0.023]$}
& \cicell{$-0.042$}{$[-0.079,-0.008]$}
& \cicell{$-0.013$}{$[-0.044,0.019]$}
& \cicell{$-0.042$}{$[-0.065,-0.019]$} \\
& \deltalabel{\opsd{} - Base}
& \cicell{$-0.067$}{$[-0.125,-0.017]$}
& \cicell{$-0.096$}{$[-0.160,-0.042]$}
& \cicell{$-0.029$}{$[-0.062,0.002]$}
& \cicell{$-0.064$}{$[-0.095,-0.037]$} \\

\bottomrule
\end{tabular*}
\end{table*}

\subsection{Full Pass@k Results}

The main text reports avg@16 values for readability. Tables in this appendix
give the corresponding full-format versions of the main result tables, with
each benchmark cell written as pass@1 / pass@16.
The pass@1 values are identical to the corresponding main-table avg@16 values,
because both are computed as mean correctness over the same 16 sampled rollouts
per problem. The pass@16 values instead estimate the probability that at least
one of 16 sampled rollouts solves the problem, using the unbiased estimator
described in Appendix~\ref{app:evaluation}.

\begin{table*}[!htbp]
\centering
\caption{\textbf{Full pass@1/pass@16 results for the Countdown-trained think-vs-instruct comparison.}
This is the full-format companion to Table~\ref{tab:countdown_think_vs_instruct}. Entries report pass@1 / pass@16 on held-out Countdown data, AIME24, AIME25, and HMMT25. The Average column averages the four benchmarks.}
\label{tab:countdown_think_vs_instruct_full}
\footnotesize
\setlength{\tabcolsep}{4pt}
\renewcommand{\arraystretch}{1.12}

\newcommand{\metrichead}[1]{%
  \textbf{#1}%
}

\begin{tabular}{@{}llccccc@{}}
\toprule
\textbf{Model} &
& \metrichead{Countdown}
& \metrichead{AIME24}
& \metrichead{AIME25}
& \metrichead{HMMT25}
& \metrichead{Average} \\
\midrule

\multirow{2}{*}{\textbf{\qwenfourbinstruct{}}}
& \textbf{Base}
& 0.841 / 0.966
& \textbf{0.604} / 0.833
& 0.463 / 0.733
& \textbf{0.304} / \textbf{0.500}
& 0.553 / 0.758 \\
&
\textbf{+\opsd{}}
& \textbf{0.865} / \textbf{0.968}
& 0.594 / \textbf{0.867}
& \textbf{0.483} / \textbf{0.767}
& 0.294 / \textbf{0.500}
& \textbf{0.559} / \textbf{0.775} \\
\midrule

\multirow{2}{*}{\textbf{\qwenfourbthink{}}}
& \textbf{Base}
& 0.945 / \textbf{0.996}
& \textbf{0.804} / \textbf{0.933}
& \textbf{0.804} / \textbf{0.900}
& \textbf{0.552} / \textbf{0.767}
& \textbf{0.776} / \textbf{0.899} \\
&
\textbf{+\opsd{}}
& \textbf{0.947} / \textbf{0.996}
& 0.800 / \textbf{0.933}
& 0.775 / \textbf{0.900}
& 0.537 / \textbf{0.767}
& 0.765 / \textbf{0.899} \\
\midrule

\multirow{2}{*}{\textbf{\olmosevenbinstruct{}}}
& \textbf{Base}
& 0.775 / 0.956
& \textbf{0.525} / \textbf{0.867}
& \textbf{0.415} / 0.700
& 0.237 / 0.467
& 0.488 / 0.747 \\
&
\textbf{+\opsd{}}
& \textbf{0.814} / \textbf{0.978}
& 0.510 / \textbf{0.867}
& 0.394 / \textbf{0.733}
& \textbf{0.256} / \textbf{0.500}
& \textbf{0.494} / \textbf{0.769} \\
\midrule

\multirow{2}{*}{\textbf{\olmosevenbthink{}}}
& \textbf{Base}
& 0.879 / 0.996
& 0.719 / 0.900
& 0.667 / 0.833
& \textbf{0.452} / \textbf{0.800}
& 0.679 / 0.882 \\
&
\textbf{+\opsd{}}
& \textbf{0.890} / \textbf{0.998}
& \textbf{0.742} / \textbf{0.933}
& \textbf{0.698} / \textbf{0.867}
& \textbf{0.452} / 0.733
& \textbf{0.695} / \textbf{0.883} \\

\bottomrule
\end{tabular}
\end{table*}

\begin{table*}[!htbp]
\centering
\caption{\textbf{Full pass@1/pass@16 results for OpenThoughts-trained thinking models.}
This is the full-format companion to Table~\ref{tab:openthoughts_thinking_models}. Entries report pass@1 / pass@16 on AIME24, AIME25, and HMMT25. The Average column averages the three benchmarks.}
\label{tab:openthoughts_thinking_models_full}
\footnotesize
\setlength{\tabcolsep}{3pt}
\renewcommand{\arraystretch}{1.12}

\newcommand{\metrichead}[1]{%
  \textbf{#1}%
}

\begin{tabular*}{\textwidth}{@{\extracolsep{\fill}}llcccc@{}}
\toprule
\textbf{Model}
&
& \metrichead{AIME24}
& \metrichead{AIME25}
& \metrichead{HMMT25}
& \metrichead{Average} \\
\midrule

\multirow{2}{*}{\textbf{\qwenonesevenb{} (Thinking)}}
& \textbf{Base}
& \textbf{0.502} / \textbf{0.800}
& \textbf{0.398} / \textbf{0.667}
& \textbf{0.215} / \textbf{0.467}
& \textbf{0.372} / \textbf{0.644} \\
&
\textbf{+\opsd{}}
& 0.435 / \textbf{0.800}
& 0.302 / 0.600
& 0.185 / 0.433
& 0.308 / 0.611 \\
\midrule

\multirow{2}{*}{\textbf{\qwenfourb{} (Thinking)}}
& \textbf{Base}
& \textbf{0.727} / \textbf{0.867}
& \textbf{0.635} / \textbf{0.867}
& \textbf{0.410} / \textbf{0.667}
& \textbf{0.591} / \textbf{0.800} \\
&
\textbf{+\opsd{}}
& 0.683 / 0.833
& 0.556 / 0.833
& 0.362 / 0.600
& 0.534 / 0.756 \\
\midrule

\multirow{2}{*}{\textbf{\qweneightb{} (Thinking)}}
& \textbf{Base}
& \textbf{0.758} / 0.833
& \textbf{0.700} / \textbf{0.833}
& \textbf{0.448} / \textbf{0.733}
& \textbf{0.635} / \textbf{0.800} \\
&
\textbf{+\opsd{}}
& 0.721 / \textbf{0.900}
& 0.613 / 0.833
& 0.400 / 0.633
& 0.578 / 0.789 \\
\midrule

\multirow{2}{*}{\textbf{\qwenfourbthink{}}}
& \textbf{Base}
& \textbf{0.804} / \textbf{0.933}
& \textbf{0.804} / 0.900
& \textbf{0.552} / 0.767
& \textbf{0.720} / \textbf{0.867} \\
&
\textbf{+\opsd{}}
& 0.787 / 0.867
& 0.731 / \textbf{0.900}
& 0.529 / \textbf{0.800}
& 0.683 / 0.856 \\
\midrule

\multirow{2}{*}{\textbf{\olmosevenbthink{}}}
& \textbf{Base}
& \textbf{0.719} / 0.900
& \textbf{0.667} / 0.833
& \textbf{0.452} / \textbf{0.800}
& \textbf{0.612} / \textbf{0.844} \\
&
\textbf{+\opsd{}}
& 0.715 / \textbf{0.933}
& 0.652 / \textbf{0.867}
& 0.446 / 0.667
& 0.604 / 0.822 \\

\bottomrule
\end{tabular*}
\end{table*}

\begin{table*}[!htbp]
\centering
\caption{\textbf{Full pass@1/pass@16 results for the \qwenonesevenb{} OPD comparison.}
This is the full-format companion to Table~\ref{tab:qwen17_openthoughts_opsd_opd}. Entries report pass@1 / pass@16 on AIME24, AIME25, and HMMT25. The Average column averages the three benchmarks.}
\label{tab:qwen17_openthoughts_opsd_opd_full}
\footnotesize
\setlength{\tabcolsep}{3pt}
\renewcommand{\arraystretch}{1.12}

\newcommand{\metrichead}[1]{%
  \textbf{#1}%
}

\begin{tabular*}{\textwidth}{@{\extracolsep{\fill}}llcccc@{}}
\toprule
\textbf{Model}
&
& \metrichead{AIME24}
& \metrichead{AIME25}
& \metrichead{HMMT25}
& \metrichead{Average} \\
\midrule

\multirow{4}{*}{\textbf{\qwenonesevenb{}}}
& \textbf{Base}
& 0.502 / \textbf{0.800}
& \textbf{0.398} / \textbf{0.667}
& 0.215 / 0.467
& 0.372 / 0.644 \\

&
\textbf{+\opsd{}}
& 0.435 / \textbf{0.800}
& 0.302 / 0.600
& 0.185 / 0.433
& 0.308 / 0.611 \\

&
\textbf{+OPD}
& \textbf{0.540} / \textbf{0.800}
& 0.385 / \textbf{0.667}
& \textbf{0.252} / 0.567
& \textbf{0.392} / 0.678 \\

&
\textbf{+OPD gold demo}
& 0.467 / \textbf{0.800}
& 0.344 / \textbf{0.667}
& 0.240 / \textbf{0.600}
& 0.350 / \textbf{0.689} \\

\bottomrule
\end{tabular*}
\end{table*}

\subsection{OPD Accuracy, Coverage, and Answer Diversity}

Average accuracy and pass@$k$ measure different properties. On these 30-problem
benchmarks, one additional problem solved changes the benchmark-level pass@16
value by $0.033$. We therefore supplement the pass@$k$ results with problem-level
coverage and answer-diversity statistics for OPD and gold-demonstration OPD.

\begin{table}[t]
\centering
\caption{\textbf{Gold-demonstration OPD increases answer diversity without a
statistically detectable increase in problem coverage.} A problem is counted
as solved when at least one of its 16 sampled responses is correct. Diversity
statistics are averaged across problems.}
\label{tab:opd_coverage_diversity}
\small
\begin{tabular}{lcc}
\toprule
\textbf{Metric} & \textbf{+OPD} & \textbf{+OPD gold demo} \\
\midrule
AIME24 problems solved & 24/30 & 24/30 \\
AIME25 problems solved & 20/30 & 20/30 \\
HMMT25 problems solved & 18/30 & 17/30 \\
Distinct boxed answers per problem & 5.98 & 6.68 \\
Answer entropy & 1.78 bits & 1.96 bits \\
\bottomrule
\end{tabular}
\end{table}

An exact paired McNemar test over all 90 problems gives $p=1.00$. The
paired-bootstrap difference between gold-demonstration OPD and OPD at $k=1$ is
$-0.042$, with a 95\% confidence interval of $[-0.066,-0.019]$. The difference
remains statistically significant at $k=2$ but is indistinguishable from zero
by $k=8$. Although gold-demonstration OPD produces more diverse answers, this
diversity does not yield a statistically detectable increase in the number of
problems solved.

These results support a reduction in single-sample accuracy at $k=1$ and $k=2$,
but do not demonstrate a reduction in pass@16 or search coverage. The pass@16
result neither supports nor contradicts the single-sample accuracy result.

\clearpage
\section{Multi-Seed Robustness Analysis}
\label{app:statistical-robustness}

\subsection{Training-Seed Robustness}
\label{app:training-seed-robustness}

The main OpenThoughts comparison uses the seed-42 runs. To test whether the
observed degradation is specific to those training runs, we additionally train
each \opsd{} model with seeds 43 and 44 under the same training and evaluation
recipe. Table~\ref{tab:openthoughts_seed_robustness} reports every seed and the
mean $\pm$ standard deviation across the three seeds. All 15 \opsd{} runs score
below their corresponding base model, and the standard deviation of the
three-seed average is at most $0.009$ across models.

The final column reports 95\% confidence intervals for the average \opsd{}
minus base delta, using the paired problem-level bootstrap protocol described
in Appendix~\ref{app:bootstrap-cis} and pooling the three seed-specific \opsd{}
evaluations. These intervals capture both the replicated training runs and
evaluation-problem variation. They exclude zero for all four Qwen models. The
OLMo comparison has the same negative direction, but its interval includes
zero.

\begin{table}[!htbp]
\centering
\caption{\textbf{The OpenThoughts degradation is robust across training seeds.}
We report avg@16 for the base model, each \opsd{} training seed, and the
three-seed mean $\pm$ standard deviation. The final column gives the pooled
paired problem-level bootstrap interval for the average \opsd{} minus base
delta across AIME24, AIME25, and HMMT25.}
\label{tab:openthoughts_seed_robustness}
\scriptsize
\setlength{\tabcolsep}{3.5pt}
\renewcommand{\arraystretch}{1.08}
\resizebox{\textwidth}{!}{%
\begin{tabular}{llccccc}
\toprule
\textbf{Model} & & \textbf{AIME24} & \textbf{AIME25} & \textbf{HMMT25} & \textbf{Average} & \textbf{Pooled $\Delta$ [95\% CI]} \\
\midrule
\multirow{5}{*}{\textbf{Qwen3-1.7B (Thinking)}} & \textbf{Base} & $\mathbf{0.502}\phantom{{}\pm 0.000}$ & $\mathbf{0.398}\phantom{{}\pm 0.000}$ & $\mathbf{0.215}\phantom{{}\pm 0.000}$ & $\mathbf{0.372}\phantom{{}\pm 0.000}$ & \\
 & +OPSD (seed 42) & $0.435\phantom{{}\pm 0.000}$ & $0.302\phantom{{}\pm 0.000}$ & $0.185\phantom{{}\pm 0.000}$ & $0.308\phantom{{}\pm 0.000}$ & \\
 & +OPSD (seed 43) & $0.427\phantom{{}\pm 0.000}$ & $0.329\phantom{{}\pm 0.000}$ & $0.215\phantom{{}\pm 0.000}$ & $0.324\phantom{{}\pm 0.000}$ & \\
 & +OPSD (seed 44) & $0.408\phantom{{}\pm 0.000}$ & $0.327\phantom{{}\pm 0.000}$ & $0.192\phantom{{}\pm 0.000}$ & $0.309\phantom{{}\pm 0.000}$ & \\
 & +OPSD (mean) & $0.424 \pm 0.014$ & $0.319 \pm 0.015$ & $0.197 \pm 0.015$ & $0.313 \pm 0.009$ & $-0.058\;[-0.084,-0.034]$ \\
\midrule
\multirow{5}{*}{\textbf{Qwen3-4B (Thinking)}} & \textbf{Base} & $\mathbf{0.727}\phantom{{}\pm 0.000}$ & $\mathbf{0.635}\phantom{{}\pm 0.000}$ & $\mathbf{0.410}\phantom{{}\pm 0.000}$ & $\mathbf{0.591}\phantom{{}\pm 0.000}$ & \\
 & +OPSD (seed 42) & $0.683\phantom{{}\pm 0.000}$ & $0.556\phantom{{}\pm 0.000}$ & $0.362\phantom{{}\pm 0.000}$ & $0.534\phantom{{}\pm 0.000}$ & \\
 & +OPSD (seed 43) & $0.679\phantom{{}\pm 0.000}$ & $0.554\phantom{{}\pm 0.000}$ & $0.365\phantom{{}\pm 0.000}$ & $0.533\phantom{{}\pm 0.000}$ & \\
 & +OPSD (seed 44) & $0.679\phantom{{}\pm 0.000}$ & $0.556\phantom{{}\pm 0.000}$ & $0.388\phantom{{}\pm 0.000}$ & $0.541\phantom{{}\pm 0.000}$ & \\
 & +OPSD (mean) & $0.681 \pm 0.002$ & $0.556 \pm 0.001$ & $0.372 \pm 0.014$ & $0.536 \pm 0.004$ & $-0.055\;[-0.081,-0.031]$ \\
\midrule
\multirow{5}{*}{\textbf{Qwen3-8B (Thinking)}} & \textbf{Base} & $\mathbf{0.758}\phantom{{}\pm 0.000}$ & $\mathbf{0.700}\phantom{{}\pm 0.000}$ & $\mathbf{0.448}\phantom{{}\pm 0.000}$ & $\mathbf{0.635}\phantom{{}\pm 0.000}$ & \\
 & +OPSD (seed 42) & $0.721\phantom{{}\pm 0.000}$ & $0.613\phantom{{}\pm 0.000}$ & $0.400\phantom{{}\pm 0.000}$ & $0.578\phantom{{}\pm 0.000}$ & \\
 & +OPSD (seed 43) & $0.735\phantom{{}\pm 0.000}$ & $0.617\phantom{{}\pm 0.000}$ & $0.412\phantom{{}\pm 0.000}$ & $0.588\phantom{{}\pm 0.000}$ & \\
 & +OPSD (seed 44) & $0.702\phantom{{}\pm 0.000}$ & $0.606\phantom{{}\pm 0.000}$ & $0.400\phantom{{}\pm 0.000}$ & $0.569\phantom{{}\pm 0.000}$ & \\
 & +OPSD (mean) & $0.719 \pm 0.017$ & $0.612 \pm 0.005$ & $0.404 \pm 0.007$ & $0.578 \pm 0.009$ & $-0.057\;[-0.084,-0.033]$ \\
\midrule
\multirow{5}{*}{\textbf{Qwen3-4B-Think-2507}} & \textbf{Base} & $\mathbf{0.804}\phantom{{}\pm 0.000}$ & $\mathbf{0.804}\phantom{{}\pm 0.000}$ & $\mathbf{0.552}\phantom{{}\pm 0.000}$ & $\mathbf{0.720}\phantom{{}\pm 0.000}$ & \\
 & +OPSD (seed 42) & $0.787\phantom{{}\pm 0.000}$ & $0.731\phantom{{}\pm 0.000}$ & $0.529\phantom{{}\pm 0.000}$ & $0.683\phantom{{}\pm 0.000}$ & \\
 & +OPSD (seed 43) & $0.773\phantom{{}\pm 0.000}$ & $0.729\phantom{{}\pm 0.000}$ & $0.523\phantom{{}\pm 0.000}$ & $0.675\phantom{{}\pm 0.000}$ & \\
 & +OPSD (seed 44) & $0.790\phantom{{}\pm 0.000}$ & $0.746\phantom{{}\pm 0.000}$ & $0.523\phantom{{}\pm 0.000}$ & $0.686\phantom{{}\pm 0.000}$ & \\
 & +OPSD (mean) & $0.783 \pm 0.009$ & $0.735 \pm 0.009$ & $0.525 \pm 0.004$ & $0.681 \pm 0.006$ & $-0.039\;[-0.064,-0.013]$ \\
\midrule
\multirow{5}{*}{\textbf{OLMo-7B-Think}} & \textbf{Base} & $\mathbf{0.719}\phantom{{}\pm 0.000}$ & $\mathbf{0.667}\phantom{{}\pm 0.000}$ & $\mathbf{0.452}\phantom{{}\pm 0.000}$ & $\mathbf{0.612}\phantom{{}\pm 0.000}$ & \\
 & +OPSD (seed 42) & $0.715\phantom{{}\pm 0.000}$ & $0.652\phantom{{}\pm 0.000}$ & $0.446\phantom{{}\pm 0.000}$ & $0.604\phantom{{}\pm 0.000}$ & \\
 & +OPSD (seed 43) & $0.715\phantom{{}\pm 0.000}$ & $0.644\phantom{{}\pm 0.000}$ & $0.425\phantom{{}\pm 0.000}$ & $0.594\phantom{{}\pm 0.000}$ & \\
 & +OPSD (seed 44) & $0.713\phantom{{}\pm 0.000}$ & $0.650\phantom{{}\pm 0.000}$ & $0.425\phantom{{}\pm 0.000}$ & $0.596\phantom{{}\pm 0.000}$ & \\
 & +OPSD (mean) & $0.714 \pm 0.001$ & $0.649 \pm 0.004$ & $0.432 \pm 0.012$ & $0.598 \pm 0.005$ & $-0.014\;[-0.031,0.002]$ \\
\bottomrule
\end{tabular}%
}
\end{table}

\clearpage
\section{Fork/Lock Token Measurement}
\label{app:fork-lock-measurement}

For each model family, we measured fork- and lock-like token positions by
evaluating teacher next-token distributions on fixed student traces. We used the
same 60 OpenMathReasoning prompts for every model and generated one student
reasoning trace per prompt using the base student prompt. For each trace, we
evaluated the teacher distribution at every generated token position under three
conditioning settings: \emph{base}, \emph{sparse}, and \emph{dense}. The base
condition included only the problem statement; the sparse condition additionally
provided the correct final answer; and the dense condition provided privileged
context in the form of a truncated reference solution from a stronger model.

For each token position $t$, we formed the teacher context $(x, y_{<t})$ and
stored the teacher's retained top-$K$ next-token log-probabilities. In the final
SRT runs, we used $K=3$ and set \texttt{max\_student\_tokens} to 3072 for the
base and sparse conditions. For dense runs, we capped the reference trace at
2048 tokens and the student trace at 1536 tokens to avoid prompt-logprob memory
failures. All six cells per model were completed: OPSD and OPD crossed with
base, sparse, and dense, with 60 traces per cell.

\paragraph{Entropy-threshold analysis.}
We first normalized the entropy of the retained top-$K$ distribution:
\[
H_K^{\mathrm{norm}}
=
\frac{-\sum_{i=1}^{K} q_i \log q_i}{\log K},
\]
where $q_i$ denotes the top-$K$ probabilities renormalized over the retained
support. Positions with $H_K^{\mathrm{norm}} \le 0.20$ were labeled \emph{lock}
tokens, positions with $H_K^{\mathrm{norm}} \ge 0.60$ were labeled \emph{fork}
tokens, and all remaining positions were labeled neutral.

\paragraph{Support-aware SSD approximation.}
We also classified positions using the geometry of the retained support.
Starting from the saved top-$K$ distribution, we applied top-$p$ truncation with
$p=0.8$ and computed the retained support size, top-token probability,
top-1/top-2 log-probability gap, entropy-derived effective support size
\[
N_{\mathrm{eff}} = \exp(H_S),
\]
and the number of competitive tokens within a factor of 3 of the top token. A
position was labeled lock-like when the retained support was sharply
concentrated, and fork-like when multiple retained tokens remained competitive.
Tokens outside the retained support were treated as tail mass rather than forks.
Positions satisfying neither criterion were labeled neutral.

\paragraph{Aggregation.}
For each trace and conditioning setting, we computed fork, lock, and neutral
rates as the fraction of classified token positions in the trace. We visualize
per-trace rates using boxplots, separately for OPSD and OPD, with the base,
sparse, and dense conditions shown in each panel. Boxes summarize the
distribution across 60 traces; jittered points show individual traces; and
diamond markers indicate means.

\clearpage
\section{OPD Ablations}
\label{app:opd-ablations}

We ablate where the OPD loss is applied in the \qwenonesevenb{} OpenThoughts
comparison from Table~\ref{tab:qwen17_openthoughts_opsd_opd}. Vanilla OPD
applies the unprivileged teacher's loss to all sampled response tokens.
Epistemic-token OPD applies the same loss only to tokens in the
epistemic-marker set. Random-fraction OPD is a token-count-matched control: if
$x$ is the average fraction of epistemic tokens in student responses, then each
rollout receives OPD loss on a uniformly sampled $x\%$ subset of response
tokens. OPD + privileged gold-demonstration context uses the same loss over
response tokens but conditions the teacher on a gold demonstration.

\begin{table*}[!htbp]
\centering
\caption{\textbf{Token-masked OPD ablations do not reproduce the gold-demonstration degradation pattern.}
We evaluate \qwenonesevenb{} OPD variants on AIME24, AIME25, and HMMT25. Entries report pass@1 / pass@16 to match the full-format tables in Appendix~\ref{app:full-passk-results}. Epistemic-token OPD applies the loss only on epistemic-marker tokens. Random-fraction OPD applies the loss to a random fraction of response tokens matched to the average epistemic-token rate. Both token-masked OPD variants improve over the base model on average, while the gold-demonstration variant drops below the base on average at pass@1.}
\label{tab:opd_ablation_passk}
\footnotesize
\setlength{\tabcolsep}{3pt}
\renewcommand{\arraystretch}{1.12}

\newcommand{\metrichead}[1]{%
  \textbf{#1}%
}

\begin{tabular*}{\textwidth}{@{\extracolsep{\fill}}llcccc@{}}
\toprule
\textbf{Model}
&
& \metrichead{AIME24}
& \metrichead{AIME25}
& \metrichead{HMMT25}
& \metrichead{Average} \\
\midrule
\multirow{5}{*}{\textbf{\qwenonesevenb{}}}
& \textbf{Base}
& 0.502 / \textbf{0.800}
& 0.398 / 0.667
& 0.215 / 0.467
& 0.372 / 0.644 \\
&
\textbf{Vanilla OPD}
& \textbf{0.540} / \textbf{0.800}
& 0.385 / 0.667
& 0.252 / 0.567
& \textbf{0.392} / 0.678 \\
&
\textbf{Epistemic-only OPD}
& 0.523 / \textbf{0.800}
& 0.396 / 0.733
& 0.235 / 0.533
& 0.385 / 0.689 \\
&
\textbf{Random-frac OPD}
& 0.494 / \textbf{0.800}
& \textbf{0.408} / \textbf{0.767}
& \textbf{0.256} / 0.533
& 0.386 / \textbf{0.700} \\
&
\textbf{OPD + gold demo}
& 0.467 / \textbf{0.800}
& 0.344 / 0.667
& 0.240 / \textbf{0.600}
& 0.350 / 0.689 \\
\bottomrule
\end{tabular*}
\end{table*}

Table~\ref{tab:opd_ablation_passk} suggests that token-masked OPD can still
offer some improvement over the base thinking model even when the loss is
applied to only a small fraction of response tokens. The epistemic-token and
random matched-fraction masks are close enough that these results do not clearly
rank one mask above the other. This is consistent with the idea that lexical
epistemic markers such as \textit{wait} and \textit{hmm} are useful proxies for
forking behavior, but do not exhaust it: branch-relevant decisions can also
occur on ordinary mathematical, connective, or formatting tokens. A random
matched-fraction mask may therefore sample some consequential non-lexical
decision points, while the epistemic-token mask targets explicit deliberation
markers more directly.

\begin{table*}[t]
\centering
\caption{\textbf{Gold-demonstration context also lowers probability mass on epistemic tokens.}
This token-level companion to Table~\ref{tab:qwen17_openthoughts_epistemic_tokens} reports the model probability assigned to epistemic markers. The Marginal column gives aggregate probability mass on the epistemic-token set; the named columns give log-probabilities for representative markers; and Avg. logp averages over the set. Vanilla OPD leaves these probabilities nearly unchanged, while the gold-demonstration variant lowers both the aggregate marginal and several revision-token log-probabilities.}
\label{tab:qwen17_openthoughts_epistemic_logprobs}
\scriptsize
\setlength{\tabcolsep}{2pt}
\renewcommand{\arraystretch}{1.12}

\begin{tabular*}{\textwidth}{@{\extracolsep{\fill}}lccccccccc@{}}
\toprule
\textbf{Method}
& \textbf{Marginal}
& \textbf{wait}
& \textbf{recall}
& \textbf{okay}
& \textbf{altern}
& \textbf{check}
& \textbf{verify}
& \textbf{hmm}
& \textbf{Avg. logp} \\
\midrule
\textbf{Base}
& 0.00928
& -0.57
& -0.39
& -0.15
& -0.64
& -0.38
& -0.66
& -0.72
& -0.502 \\
\textbf{+OPD}
& 0.00929
& -0.57
& -0.38
& -0.15
& -0.64
& -0.38
& -0.67
& -0.70
& -0.499 \\
\textbf{+OPD gold demo}
& 0.00853
& -0.72
& -0.50
& -0.17
& -0.82
& -0.40
& -0.67
& -0.96
& -0.605 \\
\bottomrule
\end{tabular*}
\end{table*}

\begin{table*}[!htbp]
\centering
\caption{\textbf{Sparse-loss OPD controls leave epistemic-marker probabilities close to vanilla OPD.}
We report aggregate probability mass on the epistemic-marker set and log-probabilities for representative markers. Epistemic-token OPD and random-fraction OPD remain nearly identical to vanilla OPD on the aggregate marginal and average log-probability. Conditioning the teacher on a privileged gold demonstration lowers the marginal mass and assigns substantially lower probability to several revision markers, especially \textit{wait}, \textit{recall}, \textit{altern}, and \textit{hmm}.}
\label{tab:opd_ablation_epistemic_logprobs}
\scriptsize
\setlength{\tabcolsep}{2pt}
\renewcommand{\arraystretch}{1.12}

\begin{tabular*}{\textwidth}{@{\extracolsep{\fill}}lccccccccc@{}}
\toprule
\textbf{Method}
& \textbf{Marginal}
& \textbf{wait}
& \textbf{recall}
& \textbf{okay}
& \textbf{altern}
& \textbf{check}
& \textbf{verify}
& \textbf{hmm}
& \textbf{Avg. logp} \\
\midrule
\textbf{Base}
& 0.00928
& -0.57
& -0.39
& -0.15
& -0.64
& -0.38
& -0.66
& -0.72
& -0.502 \\
\textbf{Vanilla OPD}
& 0.00929
& -0.57
& -0.38
& -0.15
& -0.64
& -0.38
& -0.67
& -0.70
& -0.499 \\
\textbf{Epistemic-only OPD}
& 0.00928
& -0.57
& -0.38
& -0.15
& -0.64
& -0.38
& -0.66
& -0.71
& -0.498 \\
\textbf{Random-frac OPD}
& 0.00928
& -0.57
& -0.37
& -0.15
& -0.64
& -0.37
& -0.67
& -0.72
& -0.499 \\
\textbf{OPD + gold demo}
& 0.00853
& -0.72
& -0.50
& -0.17
& -0.82
& -0.40
& -0.67
& -0.96
& -0.605 \\
\bottomrule
\end{tabular*}
\end{table*}

\begin{table*}[!htbp]
\centering
\caption{\textbf{Gold-demonstration context produces the largest drop in realized epistemic-token density.}
The aggregate density column reports the fraction of generated tokens in the epistemic-marker set. The remaining columns report occurrences per 1,000 generated tokens for representative markers. Epistemic-token OPD and random-fraction OPD slightly reduce aggregate marker density relative to the base and vanilla OPD, but the gold-demonstration variant produces the largest decrease, including clear reductions in \textit{wait} and \textit{hmm}.}
\label{tab:opd_ablation_epistemic_tokens}
\scriptsize
\setlength{\tabcolsep}{2pt}
\renewcommand{\arraystretch}{1.12}

\begin{tabular*}{\textwidth}{@{\extracolsep{\fill}}lcccccccc@{}}
\toprule
\textbf{Method}
& \shortstack{\textbf{Epistemic token}\\[-1pt]\textbf{density}}
& \textbf{wait}
& \textbf{recall}
& \textbf{okay}
& \textbf{altern}
& \textbf{check}
& \textbf{verify}
& \textbf{hmm} \\
\midrule
\textbf{Base}
& 1.080\%
& 3.85
& 1.11
& 2.17
& 0.64
& 1.33
& 0.25
& 1.45 \\
\textbf{Vanilla OPD}
& 1.074\%
& 3.79
& 1.04
& 2.17
& 0.72
& 1.19
& 0.16
& 1.68 \\
\textbf{Epistemic-only OPD}
& 1.047\%
& 3.22
& 1.05
& 2.23
& 0.82
& 1.43
& 0.18
& 1.54 \\
\textbf{Random-frac OPD}
& 1.039\%
& 3.24
& 1.11
& 2.17
& 0.90
& 1.27
& 0.18
& 1.52 \\
\textbf{OPD + gold demo}
& 0.850\%
& 2.54
& 0.96
& 2.07
& 0.66
& 1.05
& 0.10
& 1.11 \\
\bottomrule
\end{tabular*}
\end{table*}

\clearpage
\section{SD-Zero Self-Revision Pipeline}
\label{app:sd-zero-srt}

The interpretation in Section~\ref{sec:trained-student-deliberation} is also
consistent with an existing self-distillation pipeline in which the \opsd{} stage
is problematic for thinking models even when surrounding stages help. SD-Zero
\citep{he2026selfdistillationzero} first trains a model to revise its own
responses using reward feedback, then distills the reviser back into the
generator with an on-policy self-distillation step. We compare the base, the
self-revision training stage alone (SRT), and the full SRT+\opsd{} pipeline on
Qwen3-4B-Instruct and Qwen3-4B.

\begin{table}[t]
\centering
\caption{\textbf{The \opsd{} stage helps an instruction-tuned model but hurts
a thinking model.}
We compare the base model, \opsd{} alone, self-revision training alone (SRT),
and the full SRT+\opsd{} pipeline on Qwen3-4B-Instruct and Qwen3-4B.
Entries report avg@8 accuracy on AIME24, AIME25, HMMT25, and their average.}
\label{tab:sd_zero_result}
\small
\setlength{\tabcolsep}{4pt}
\renewcommand{\arraystretch}{1.12}
\begin{tabular*}{\textwidth}{@{\extracolsep{\fill}}llcccc@{}}
\toprule
\textbf{Model}
&
& \textbf{AIME24}
& \textbf{AIME25}
& \textbf{HMMT25}
& \textbf{Average} \\
\midrule

\multirow{4}{*}{\textbf{Qwen3-4B-Instruct}}
& \textbf{Base}
& 59.6
& 45.8
& 26.7
& 44.0 \\

&
\textbf{+\opsd{}}
& 63.3
& 47.9
& 32.9
& 48.0 \\

&
\textbf{SRT}
& 66.7
& 59.2
& 40.0
& 55.3 \\

&
\textbf{SRT + \opsd{}}
& \textbf{68.3}
& \textbf{60.0}
& \textbf{45.4}
& \textbf{57.9} \\

\midrule

\multirow{4}{*}{\textbf{Qwen3-4B (Thinking)}}
& \textbf{Base}
& 72.5
& \textbf{65.4}
& 45.4
& 61.1 \\

&
\textbf{+\opsd{}}
& 63.3
& 60.0
& 44.6
& 56.0 \\

&
\textbf{SRT}
& \textbf{73.3}
& 63.3
& \textbf{50.0}
& \textbf{62.2} \\

&
\textbf{SRT + \opsd{}}
& 70.0
& 63.3
& 43.3
& 58.9 \\

\bottomrule
\end{tabular*}
\end{table}

Table~\ref{tab:sd_zero_result} shows that the pipeline behaves as intended on
the instruction-tuned model: SRT improves the base by $11.3$ points
($44.0 \to 55.3$), and the \opsd{} stage adds another $2.6$ points
($55.3 \to 57.9$). On the thinking model, SRT also helps slightly
($61.1 \to 62.2$), but the subsequent \opsd{} stage reverses the gain and leaves
the model $3.3$ points below SRT alone ($62.2 \to 58.9$). The self-revision
stage is not the problem; the \opsd{} stage that follows it is.

\clearpage
\section{Additional Budget-Curve Figures}
\label{app:budget-curves}

\begin{figure}[!htbp]
    \centering
    \includegraphics[width=\linewidth]{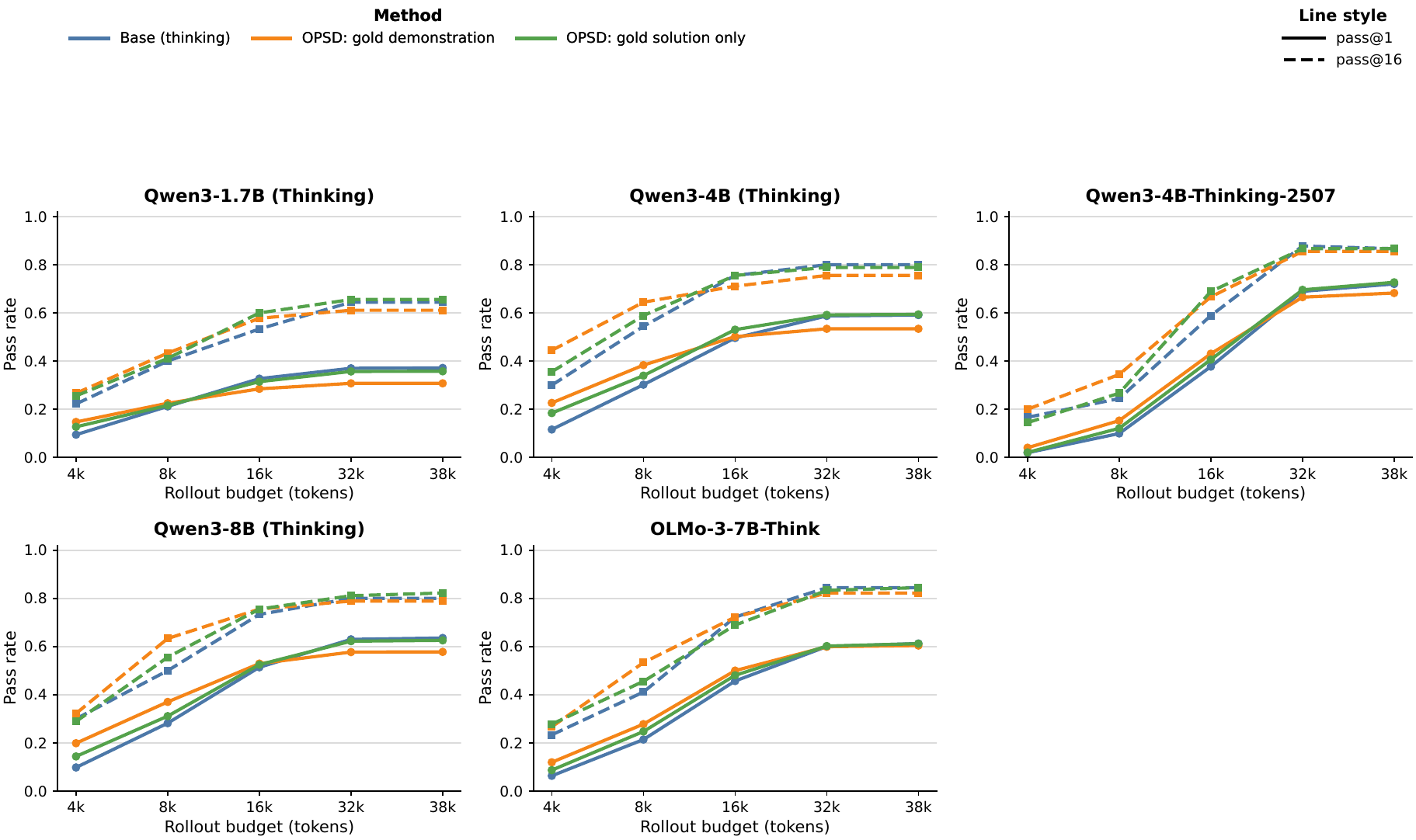}
    \caption{
    \textbf{Per-model pass-rate budget curves separate the accuracy effects in Figure~\ref{fig:openthoughts_accuracy_length_collapse}.}
    We evaluate five OpenThoughts-trained thinking models at rollout budgets from 4k to 38k tokens on AIME24, AIME25, and HMMT25.
    Solid lines show pass@1 and dashed lines show pass@16.
    Blue curves are base thinking models, orange curves are \opsd{} with full gold-demonstration context, and green curves are \opsd{} with final-answer-only privileged context.
    Dense demonstrations tend to give larger short-budget gains, while the advantage narrows or reverses at longer budgets for several models.
    }
    \label{fig:openthoughts_budget_pass_rates_thinking5_fs_ao}
\end{figure}

\begin{figure}[!htbp]
    \centering
    \includegraphics[width=\linewidth]{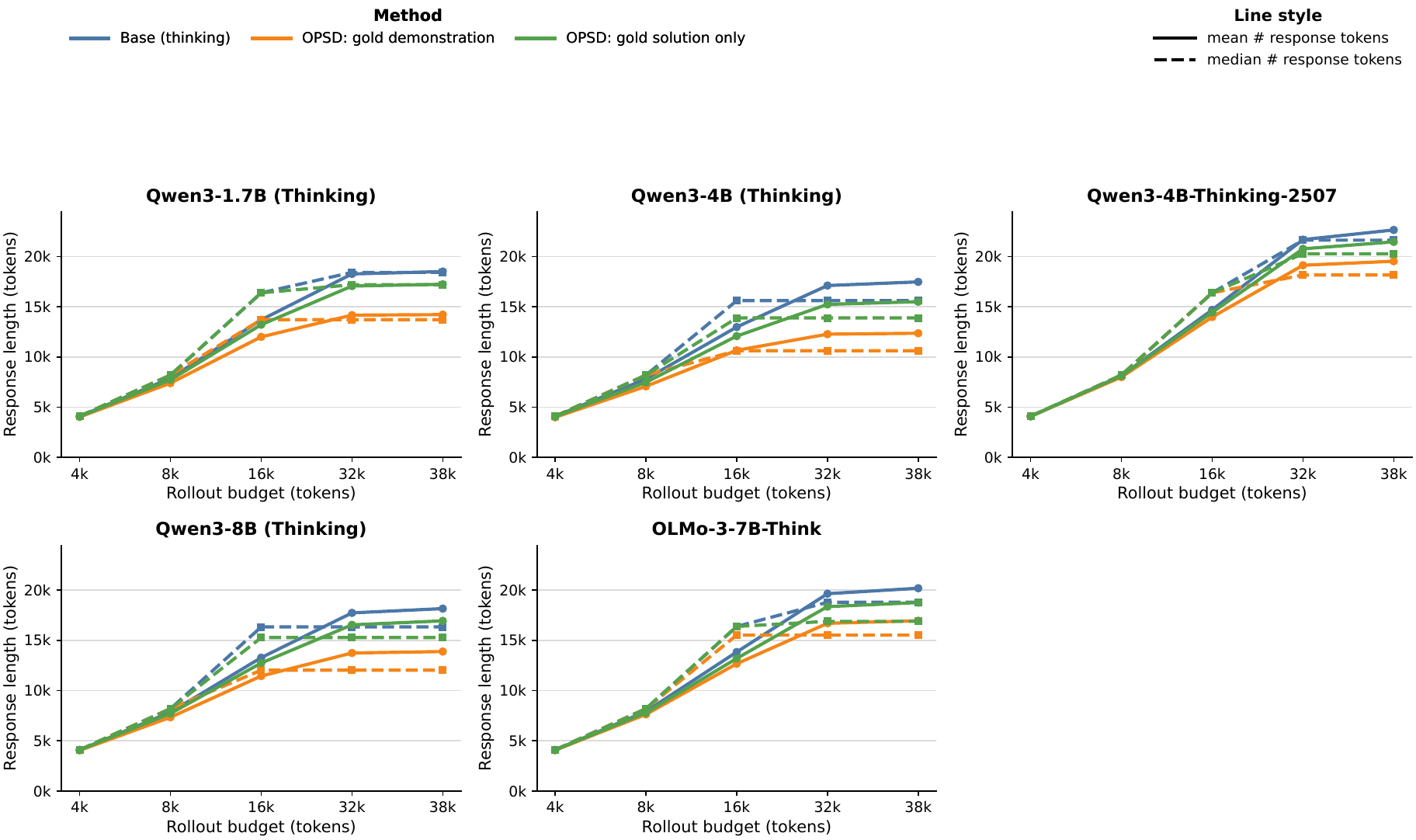}
    \caption{
    \textbf{Per-model response-length budget curves show where \opsd{} compresses long thinking rollouts.}
    This companion to Figure~\ref{fig:openthoughts_accuracy_length_collapse} reports response-token counts for the same five OpenThoughts-trained thinking models, evaluation benchmarks, and rollout budgets as Figure~\ref{fig:openthoughts_budget_pass_rates_thinking5_fs_ao}.
    Solid lines show mean response length and dashed lines show median response length.
    Blue curves are base thinking models, orange curves are \opsd{} with full gold-demonstration context, and green curves are \opsd{} with final-answer-only privileged context.
    At 32k--38k token budgets, full-demonstration \opsd{} generally produces shorter responses than the corresponding base model.
    }
    \label{fig:openthoughts_budget_response_lengths_thinking5_fs_ao}
\end{figure}

\begin{figure}[!htbp]
    \centering
    \includegraphics[width=\linewidth]{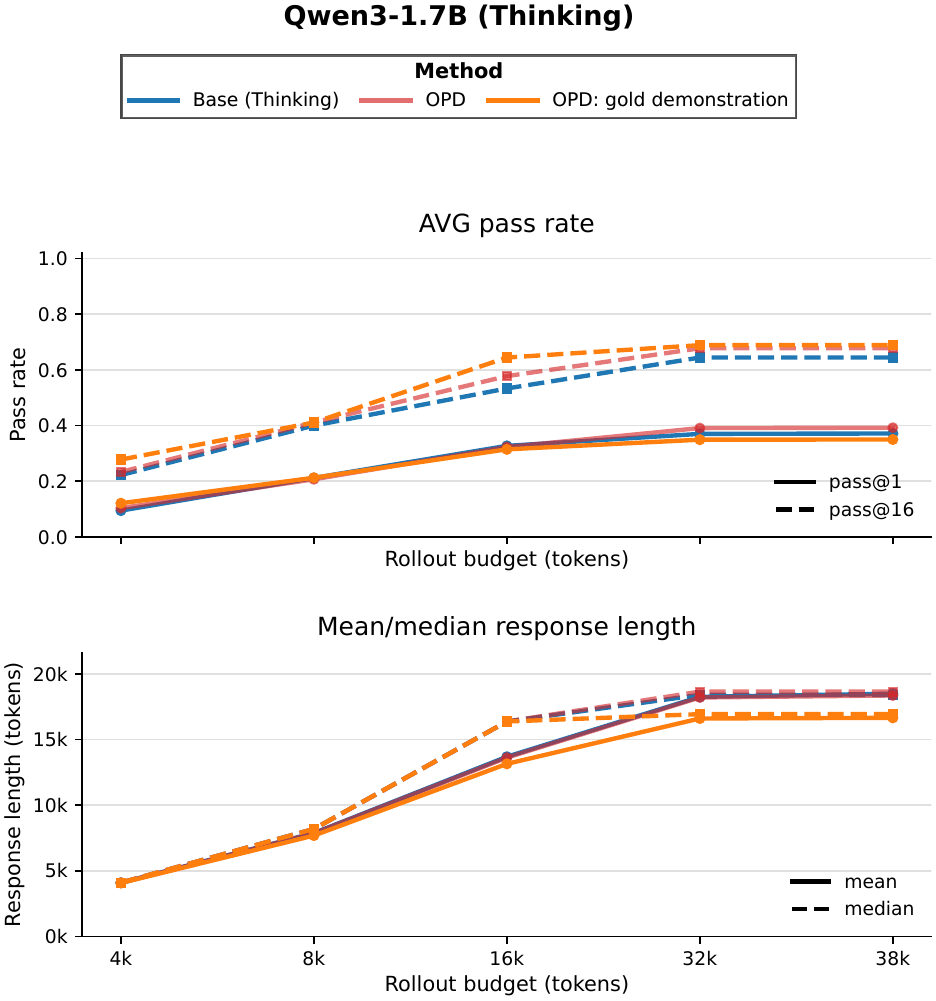}
    \caption{
    \textbf{OPD with gold-demonstration context shortens response lengths while vanilla OPD preserves the base length profile.}
    This budget-curve companion to Table~\ref{tab:qwen17_openthoughts_opsd_opd} evaluates the \qwenonesevenb{} thinking student with a larger \qweneightb{} teacher.
    We compare the base model, vanilla OPD, and OPD with gold-demonstration teacher context.
    The OPD variants are trained with a 4,096-token completion cap, and all methods are evaluated at generation caps from 4,096 to 38,912 tokens.
    Top row: pass@1 and pass@16, averaged over AIME24, AIME25, and HMMT25.
    Bottom row: mean and median response length.
    Vanilla OPD mildly improves pass@k while largely preserving the base model's response-length curve.
    Adding gold-demonstration context to the teacher shortens responses and reduces the pass@1 gains; its long-budget degradation is milder than in the \opsd{} setting, consistent with the use of a stronger larger teacher.
    }
    \label{fig:qwen17_opd_gold_demo_budget_curves}
\end{figure}

\begin{figure}[!htbp]
    \centering
    \includegraphics[width=\linewidth]{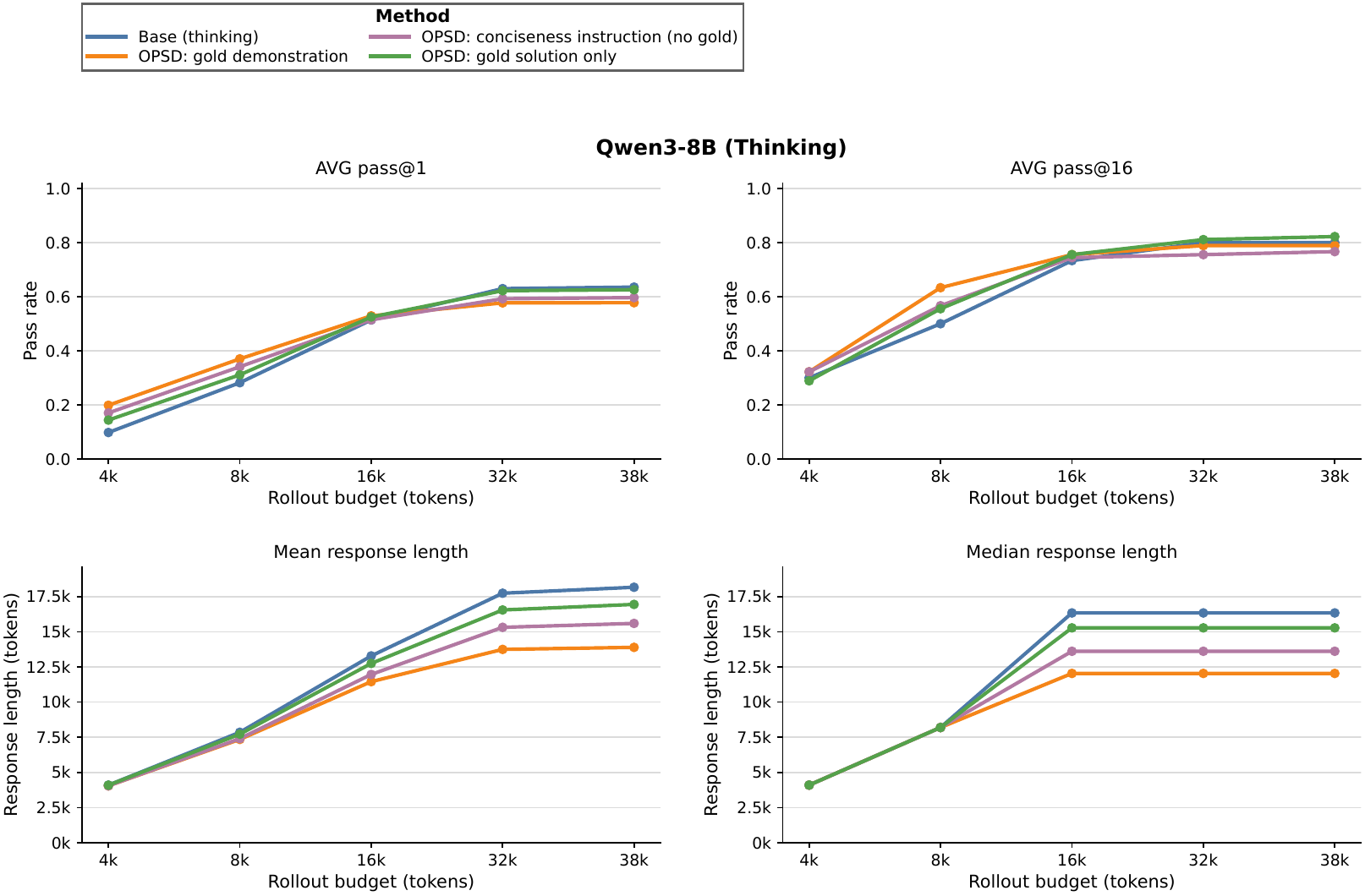}
    \caption{
    \textbf{A CRISP-style conciseness prompt compresses Qwen3-8B responses but does not recover the long-budget gains.}
    We compare base \qweneightb{} thinking, \opsd{} with full gold demonstrations, \opsd{} with final-answer-only privileged context, and a conciseness-instruction condition with no gold context, following the CRISP prompt direction of \citet{sang2026policy}.
    Top panels report pass@1 and pass@16 averaged over AIME24, AIME25, and HMMT25; bottom panels report mean and median response length.
    The conciseness condition shortens 32k--38k rollouts relative to the base and gold-solution-only runs but is less compressive than full gold demonstrations.
    Its accuracy follows the same tradeoff: it improves short-budget performance but, at long budgets, remains below the base and gold-solution-only curves, suggesting that making the student concise alone is not enough to preserve the gains from longer thinking rollouts.
    }
    \label{fig:openthoughts_qwen8b_concise_budget}
\end{figure}

Figures~\ref{fig:openthoughts_budget_pass_rates_thinking5_fs_ao}
and~\ref{fig:openthoughts_budget_response_lengths_thinking5_fs_ao} split the
budget-dependent results in Figure~\ref{fig:openthoughts_accuracy_length_collapse}
into pass-rate and response-length views across the five thinking models in
Table~\ref{tab:openthoughts_thinking_models}.
Figure~\ref{fig:openthoughts_qwen8b_concise_budget} adds a conciseness-prompt
comparison for \qweneightb{}.
Figure~\ref{fig:qwen17_opd_gold_demo_budget_curves} gives the corresponding
budget-curve view for the context-enhanced OPD comparison in
Table~\ref{tab:qwen17_openthoughts_opsd_opd}.

\paragraph{Relation to CRISP-style reasoning compression.}
CRISP \citep{sang2026policy} studies a complementary setting in which the
teacher is conditioned on a conciseness instruction rather than on a gold answer
or reference solution. Thus, unlike gold-context \opsd{}, CRISP does not give
the teacher task-answer information, but its token-level supervision can still
act globally across the rollout: the teacher is encouraged to prefer shorter,
more direct continuations at many positions. Our Qwen3-8B conciseness-prompt
control confirms the first-order CRISP effect, namely that such conditioning
shortens responses. However, the comparison with final-answer-only and
full-demonstration \opsd{} shows that response shortening is not unique to
conciseness distillation. Dense gold-demonstration context produces the
strongest long-budget compression, while final-answer-only context remains
closer to the base model. Thus, our claim is not that compression itself is
always harmful, but that token-level teachers which broadly suppress
deliberative continuations can remove the long-budget gains of thinking models.

\clearpage

\end{document}